\documentclass[a4paper,man,floatsintext,apacite]{apa6} %current

\usepackage{ragged2e}% http://ctan.org/pkg/{ragged2e}
\usepackage[english]{babel}
\usepackage[utf8x]{inputenc}
\usepackage{amsmath}
\usepackage{graphicx}
\usepackage[colorinlistoftodos]{todonotes}
\usepackage{xfrac}
\usepackage{graphicx}
\usepackage{xcolor}
\usepackage{caption}
\usepackage{subcaption}
\usepackage{booktabs}
\usepackage{multirow}
\usepackage{balance}
\usepackage{apacite}
\usepackage{algorithm}
\usepackage{algorithmic}
\usepackage{changes}
\usepackage[page]{appendix}
\usepackage{arydshln}
\usepackage{verbatim}
\usepackage{enumitem}
\usepackage[page]{appendix}

\usepackage[export]{adjustbox}
\usepackage{multicol}
\usepackage{hyperref}
\usepackage{natbib}

\usepackage{amsfonts}

\makeatletter
\def\adl@drawiv#1#2#3{%
        \hskip.5\tabcolsep
        \xleaders#3{#2.5\@tempdimb #1{1}#2.5\@tempdimb}%
                #2\z@ plus1fil minus1fil\relax
        \hskip.5\tabcolsep}
\newcommand{\cdashlinelr}[1]{%
  \noalign{\vskip\aboverulesep
           \global\let\@dashdrawstore\adl@draw
           \global\let\adl@draw\adl@drawiv}
  \cdashline{#1}
  \noalign{\global\let\adl@draw\@dashdrawstore
           \vskip\belowrulesep}}
\makeatother

\title{Credit Assignment: Challenges and Opportunities in Developing Human-like AI Agents}
\shorttitle{Credit Assignment in Humans and AI}
\author{ Thuy Ngoc Nguyen, Chase McDonald and Cleotilde Gonzalez*}
\affiliation{Dynamic Decision Making Laboratory, Carnegie Mellon University \\ 5000 Forbes Ave., Pittsburgh PA 15213 USA}
\authornote{
*Corresponding author: Cleotilde Gonzalez, Social and Decision Sciences Department, Carnegie Mellon University, 5000 Forbes Ave., Pittsburgh, PA, 15213. USA. E-mail: coty@cmu.edu.

This research was supported by the Defense Advanced Research Projects Agency and was accomplished under Grant Number W911NF-20-1-0006; and by the AFRL Award FA8650-20-F-6212 and sub-award number 1990692.

All human and model simulation data, all model code, and all scripts used for the analyses presented in this manuscript are available in Open Science Framework: 
\url{https://osf.io/ue4tv/?view\_only=28204c18941042cab2555097422706b9}. \\

}

\abstract{Temporal credit assignment is the process by which a delayed outcome is associated with each of the actions in a sequence. This process is essential for learning and skill development, and it is a learning challenge for natural and artificial intelligence. Computational approaches propose mechanisms to address delayed feedback; a prominent example is the temporal difference (TD) method in reinforcement learning (e.g., Q-learning). However, it is unclear whether these mechanisms are accurate representations of the way humans handle feedback delays. Cognitive models intend to represent the mental steps by which humans solve problems and perform a number of tasks, but limited research in cognitive science has addressed the credit assignment problem in humans and cognitive models. Our research presents results from various credit assignment mechanisms implemented into a cognitive model based on a theory of decisions from experience, Instance-Based Learning Theory (IBLT). We use a goal-seeking navigation task with feedback delays to represent different levels of decision complexity. Instance-Based Learning (IBL) models simulate the process of making sequential choices with different credit assignment mechanisms, including a new IBL-TD model that combines the IBL decision mechanism with the TD approach. Results from these models are compared to the outcomes and process measures of human decisions in two experiments. From these comparisons, we conclude that: (1) An IBL model that gives equal credit assignment to all decisions is able to match human performance better than other models, including IBL-TD and Q-learning; (2) IBL-TD and Q-learning models underperform compared to humans initially, but eventually, they outperform humans; (3) humans are influenced by decision complexity, while models are not. Detailed analyses of human and model behavior provide insights into the challenges ahead for capturing human behavior, and the potential opportunities to use these models for supporting human activities in future AI systems.
}


\begin{document}

\maketitle
\justifying
\setcounter{secnumdepth}{3}

\section{Introduction}
Learning the relationship between actions and outcomes is essential to behavioral adaptation and decision making in dynamic environments \citep{GONZALEZ03}. A common and difficult learning challenge occurs when a decision maker has to make a sequence of decisions without feedback, knowing the outcome only at the end of the sequence. The canonical example of a game of chess illustrates this well: a player only receives a definite win or loss at the conclusion of a game, but they must identify how each decision throughout the game contributes to the final outcome. A value to each individual decision may only be assigned in retrospect, after learning the final result of the game; and such a learning process is important to improve future decisions. This problem, known as the \textit{temporal credit assignment}, determines how credit should be assigned to intermediate actions within a sequence \citep{minsky1961steps}. The gap between when a decision is made and when the outcome of such a decision is observed is known as ``feedback delay'' in dynamic decision making research \citep{brehmer1989feedback} and it is one of the most challenging problems in learning to improve decisions over time in dynamic situations \citep{GONZALEZ03,gonzalez2017dynamic}.

Computational sciences have proposed a number of approaches to handle delayed feedback. One of the most prominent mechanisms to address the credit assignment problem is the temporal difference (TD) learning mechanism from the reinforcement learning (RL) literature:~\citep{sutton1985temporal, sutton2018reinforcement}. According to the TD approach, an agent predicts the value of intermediate states in the absence of final feedback, and uses prediction errors over small intervals to update their future predictions. There are a number of RL algorithms that utilize and build upon TD methods~\citep{van2009theoretical,hasselt2010double,xu2018meta}, including several state-of-the-art deep RL algorithms~\citep{mnih2015human,van2016deep,hessel2018rainbow}. Recent improvements of deep RL algorithms appear to enable Artificial Intelligence (AI) agents to reach a level of human performance that has not been possible before in a range of complex decision making tasks \citep{wong2021multiagent}. However, deep RL agents often require extensive training and they appear to be less ``flexible'' (i.e., unable to adapt to novel situations) compared to humans who are able to learn many different tasks quickly and rapidly generalize knowledge from one task to another task~\citep{pouncy2021model}. Even though previous work has shown the ability of RL models to account for human behavior in some dynamic decision tasks~\citep{SIMON11,GERSHMAN17}, none of the current models can account for this human ability to adapt rapidly in situations with delayed feedback, and AI agents are often inadequate to explain and predict adaptation and learning in complex environments in the way that \textit{humans} do \citep{lake2017building, pouncy2021model}. Therefore, a concern has been raised that the advance in RL algorithms is mainly centered on solving computational problems efficiently and optimally rather than on replicating the way humans \textit{actually} learn \citep{BOTVINICK19,lake2017building}. Our goal in this paper is to expose challenges and insights in the development of \textit{human-like} AI agents that adapt and learn in dynamic decision-making situations with delayed feedback. This research relies on RL agents and cognitive models of decision making that are constructed based on Instance-Based Learning Theory (IBLT) \citep{GONZALEZ03} and on detailed analyses of human actions in the same tasks.

\subsection*{Background}
Since its origin, AI has aimed at replicating various human behaviors in a computational form so that the machine behavior would be indistinguishable from that of a human \citep{lake2017building,turing1950mind}. Building accurate replications of human decisions (i.e., a ``cognitive clone'' of a human cognitive decision process) is essential to anticipate human error and to create personalized and dynamic digital assistants, as it is recently shown in various applications of cognitive models \citep{somers2020cognitive,gonzalez2021}. However, little effort is dedicated to investigating how to build ``\textit{human-like}'' models that consider the cognitive plausibility and diversity of human behavior. Particularly, it is unclear whether existing credit assignment mechanisms (e.g., TD) replicate human behavior or underperform humans. Thus, a major challenge for research in AI is to develop systems that can replicate human behavior~\citep{lake2017building}.

Given that cognitive architectures have been developed to represent an integrated view of the cognitive capacities of the human mind \citep{anderson2004integrated, ANDERSON14}, previous research work has addressed the question of how well a model is aligned with humans in tasks involving feedback delays~\citep{walsh2011learning,walsh2014navigating}. Specifically, TD credit assignment methods have been incorporated into cognitive architectures to emulate how humans process feedback delays in sequential decision-making tasks \citep{FU06}. Other cognitive modeling research suggests that people evaluate intermediate states in terms of future rewards, as predicted by TD learning \citep{walsh2011learning}. However, the primary focus of such studies has been on the similarities between neural processes and computational mechanisms, leaving significant room for the investigation and comparison of observed human behavior and credit assignment mechanisms in sequential decision making tasks.

Derived from cognitive architectures, cognitive models of decision making have demonstrated that it is possible to represent the cognitive processes of ``human-like'' decision making in a wide diversity of tasks, using the common theoretical principles of a theory of decisions from experience, IBLT \citep{phan2021}. Instance-Based Learning (IBL) models have been used to emulate human binary choice \citep{GONZALEZ11} and decisions in more complex dynamic resource allocation tasks such as the Internet of Things \citep{somers2020cognitive} and cybersecurity \citep{gonzalez2020design}. Through the years, IBLT has emerged as a comprehensive theory of the cognitive process by which humans make decisions from experience in dynamic environments \citep{GONZALEZ03, GONZALEZ11, hertwig2015}. In IBLT, the question of temporal credit assignment is addressed through a feedback process, but the development and comparison of particular mechanisms for credit assignment that emulate human behavior in IBLT is still in the early stages of exploration \citep{NGUYEN2020ICCM}.

\subsection*{Goals and Research Approach}
\begin{figure}[!htbp]
\centering
\includegraphics[width=0.8\linewidth]{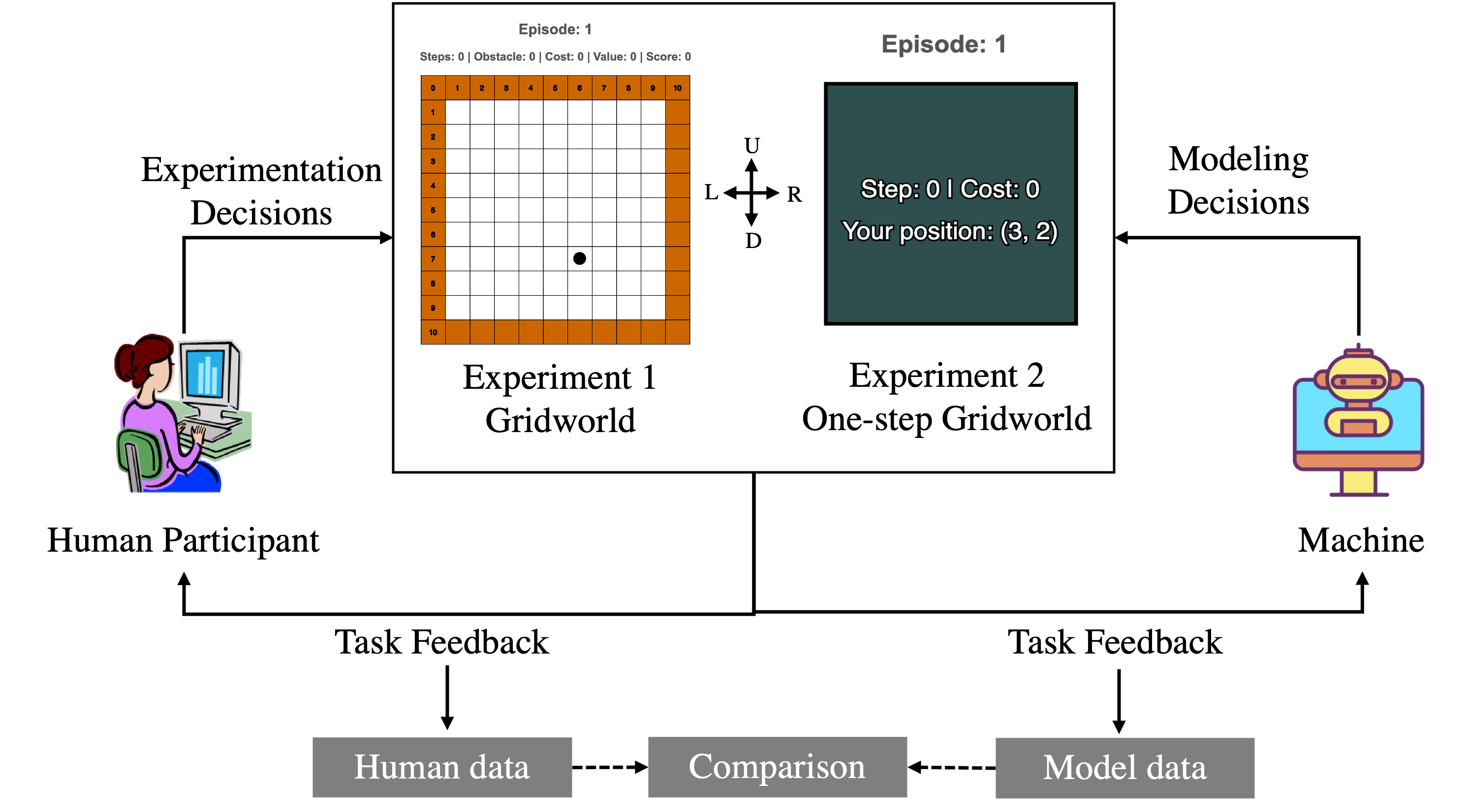}
\caption{Experimental scenarios.}
\label{fig:exp_scenario}
\end{figure}

In this research, we investigate the credit assignment problem in a goal-seeking gridworld task involving delayed feedback and various levels of decision complexity. We use an IBL model where we implement various credit assignment mechanisms and a RL model with TD, Q-learning~\citep{watkins1992q,sutton2018reinforcement}. We analyze the predictions from these models on human performance under the same gridworld tasks. We follow a general approach commonly used to validate cognitive models with human data \citep{busemeyer2010cognitive, gonzalez2017decision}, illustrated in Fig.~\ref{fig:exp_scenario}.

Our goal is to get an in-depth understanding of the ability of these models to produce \textit{human-like} behavior and to expose challenges and insights into the development of human-like AI agents. To develop insights into varying methods of temporal credit assignment, we consider three credit assignment mechanisms implemented in the IBL model: equal credit, exponential credit, and TD credit using the IBL decision mechanism. Data obtained from model simulations and human experiments are compared to determine how closely the models represent human behavior. In this research, we ran two human experiments that provided different visual representations of the same gridworld tasks. The analyses of performance and optimal actions help determine which credit assignment mechanisms produce behavior similar to humans' and which result in more optimal and effective actions than the actions of humans.  

We analyze the disparities between the models' and humans' strategies, which reveal the nuances of human behavior representations; what these AI agents can and cannot capture, and the human characteristics that may hinder optimal decisions. On the one hand, we learned that humans can create a mental model of the task given the visual representation provided, and they tend to explore and approach tasks more strategically than these models do. Our experiments reveal that humans carry concepts and strategies that influence their behavior, and that such concepts and strategies are beyond the information equivalence about the task between humans and models. The results also demonstrate that humans spend much less time exploring the environments than the models do, which results in long-term suboptimality of humans compared to TD models.

Taken together, this work brings us closer to understanding the general algorithms of credit assignment that can be used to generate human-like models. It also exposes the challenges that researchers need to address to capture the initial strategic behavior that humans might carry from a task to another; as well as potential ways to enhance human decisions in sequential decision-making tasks with delayed feedback. The findings and insights will need to be used in future work to develop human-like systems and to use these models for supporting human activities in future AI systems.

\section{Credit Assignment in Instance-Based Learning}
IBLT is a cognitive theory of decisions from experience that provides a decision making algorithm and a set of cognitive mechanisms that can be used to implement computational models of human dynamic decisions. The algorithm involves the recognition and retrieval of past experiences (i.e., instances) according to their similarity to a current decision situation, the generation of expected utility of various decision alternatives, and a choice rule that generalizes from experience \citep{GONZALEZ03}.

An ``instance'' in IBLT is a memory unit that results from the potential alternatives evaluated. These are memory representations consisting of three elements: a situation (a set of attributes that give a context to the decision, or state $S$); a decision (the action taken corresponding to an alternative in state $S$, or action $A$); and a utility (expected utility or experienced outcome $x$ of the action taken in a state).

An option $k=(S,A)$ is defined by taking action $A$ in state $S$.
At time $t$, assume that there are $n_{kt}$ different generated instances $(k,x_{ikt})$ for $i = 1,...,n_{kt}$, corresponding to selecting $k$ and achieving outcome $x_{ikt}$. Each instance $i$ in memory has an \textit{activation} value, which represents how readily available that information is in memory~\citep{ANDERSON14}, and it is determined by similarity to past situations, recency, frequency and noise~\citep{ANDERSON14}. 
Here we consider a simplified version of the Activation equation which only captures how recently and frequently instances are activated:
 \begin{equation}\label{eq:activation}
 \begin{array}{l}
     Act_{ikt} = \ln{\left(\sum\limits_{t' \in T_{ikt} }(t-t')^{-d}\right)}  + \sigma\ln{\frac{1-\xi_{ikt}}{\xi_{ikt}}},
\end{array}
 \end{equation}
where $d$ and $\sigma$ are the decay and noise parameters, respectively, and $T_{ikt}\subset \{0,...,t-1\}$ is the set of the previous timestamps in which the instance $i$ was observed. The rightmost term represents the Gaussian noise for capturing individual variation in activation, and $\xi_{ikt}$ is a random number drawn from a uniform distribution $U(0, 1)$ at each time step and for each instance and option.

Activation of an instance $i$ is used to determine the probability of retrieval of that instance from memory.
% executing action $A$ at state $S$. 
The probability of an instance $i$ is a function of its activation $Act_{ikt}$ relative to the activation of all instances:
 \begin{equation} \label{eq:retrieval_prob}
 \begin{array}{l}
     p_{ikt} = \frac{e^{\sfrac{Act_{ikt}}{\tau}}}{\sum_{j = 1}^{n_{kt}}e^{\sfrac{Act_{jkt}}{\tau}}},
     \end{array}
 \end{equation}
where $\tau$ is the Boltzmann constant (i.e., the ``temperature'') in the Boltzmann distribution.
For simplicity, $\tau$ is often defined as a function of the same $\sigma$ used in the activation equation $\tau= \sigma\sqrt{2}$.

The expected utility of option $k$ is calculated based on a mechanism called \textit{blending}~\citep{LEBIERE99} as specified in IBLT~\citep{GONZALEZ03}, using the past experienced outcomes stored in each instance. Here we employ the Blending calculation as defined for discrete choice tasks~\citep{LEJARRAGA12,GONZALEZ11}:
 \begin{equation} \label{eq:blended_value}
 \begin{array}{l}
     V_{kt} = \sum_{i=1}^{n_{kt}}p_{ikt}x_{ikt}.
 \end{array}
 \end{equation}

Essentially, the blending operation (Eq.~\ref{eq:blended_value}) is the sum of all past experienced outcomes weighted by their probability of retrieval.
% where $x_i$ is the outcome stored in an instance $i$ associated with choosing $k$; and $p_i$ is the probability of retrieving the instance $i$ from memory (Eq.~\ref{eq:retrieval_prob}). % and $n$ is the number of instances stored in memory for taking action $a$ up to the last trial.
The choice rule is to select the option that corresponds to the maximum blended value, which is stochastic based on the dynamics of the probability of retrieval of an instance, that changes according to the frequency, recency, and noise effects on the instances experienced.

%\input{3_model.tex}
% \section{Credit Assignment Mechanisms}
\subsection{Credit Assignment Mechanisms in IBL models}
IBLT suggests a feedback process that uses the outcome from the environment to update and refine the expected utility of past decisions so that updated instances are reused in future decisions \citep{GONZALEZ03,phan2021}. Initial steps were made in the early development of IBLT, however, no formalization of a credit assignment mechanism was offered by the theory, and most of the tasks that have relied on IBLT included immediate feedback \citep{GONZALEZ11}. Thus, credit assignment is an underdeveloped theoretical concept in IBLT.

Here, we introduce three alternative credit assignment mechanisms which are implemented in versions of a temporally extended IBL model. These mechanisms are based on equally- and exponentially weighted utility assignment as well as a novel incorporation of the TD error from RL models \citep{sutton2018reinforcement}. We contextualize these mechanisms in a goal-seeking gridworld task used in past research \citep{NGUYEN2020ICCM,nguyen2021theory}.

The goal-seeking task is simulated in a $11\times11$ grid that contains obstacles (black cells) and four targets, denoted by their distinct coloring (blue, green, orange, and purple) (Fig.~\ref{fig:conditions}). The obstacles are defined between two randomly sampled endpoints, and the number of obstacles randomly varies from one to three. Each target has an associated value that ranges from 0 to 1, drawn from a Dirichlet distribution. We refer to the highest value target as the \emph{preferred target} and the remaining three as \emph{distractor targets}. An agent in this task navigates through the environment by making a sequence of movements, i.e., up, down, left, right, to find the preferred target while avoiding the obstacles. In each gridworld configuration, the target values are unique, although it is always the case that the net reward of reaching the preferred target via the optimal path is greater than that of any other target.
Participants or agents are penalized for each step taken in the environment (-0.01) and for running into obstacles (-0.05).

A gridworld environment can be formalized as Markov Decision Processes (MDPs). Each MDP $\mathcal{M}$ has a state space $\mathcal{S}$, and each $(x,y)$-coordinate in the grid represents a state $S \in \mathcal{S}$. At each within-episode time step $l\in\{1,...,T\}$, an agent observes their state $S_l$, then takes an action $A_l$ from a common action space $\mathcal{A}$ (up/down/left/right) to move into state $S_{l+1}$ and observes the reward (or cost) $R_l$. By executing a policy $\pi$ in the environment $\mathcal{M}$, an agent creates a trajectory denoted by $\mathcal{T}=\{(S_l, A_l)\}_{l=1}^T$.

We distinguish between the global time $t$ in an (IBL) agent's memory and the within-episode step count $l\in\{1,...,T\}$. The global IBL timing is updated across episodes; that is, it maintains a timing index across the entirety of an agent's experience.  On the other hand, $l$ is reset at the start of each episode to track the sequence of steps within a single episode.

\subsubsection{IBL-Equal: Equal Credit}
IBL-Equal is a model based upon a simple notion of credit assignment: disseminate equal credit amongst all candidate actions in a sequence at the conclusion of the task. 
%This is specifically designed for a goal-seeking task, where $x_t=\begin{cases} R_T & \text{if outcome reached} \\ R_t & \text{otherwise} \end{cases}$, where the terminal step $T$ will denote the period in which a goal is reached.
That is, if a target is reached at step $T$, the value of the target is assigned to each instance in trajectory $\mathcal{T}=\{(S_l, A_l)\}_{l=1}^T$, i.e. $x_l = R_T$ for all $(S_l, A_l)$. If a target is not reached, the step-level costs are assigned to each instance. %\in \mathcal{C}$, where $\mathcal{C}$ is the set of instances that satisfy such a condition\footnote{In practice, our candidacy condition is to update all instances that did not result in an agent running into a wall or obstacle.}. The instances that do not satisfy the condition are updated with the reward returned at their respective step, e.g., $x_l=R_l$.

\subsubsection{IBL-Exponential: Exponentially Discounted Returns} We define a second method, IBL-Exponential, where the outcome is assigned based on discounted future returns: for all instances in $\mathcal{T}$, the $l$th instance is assigned outcome $x_l=\gamma^{T-l} R_{T}$ , where $\gamma$ is a temporal discount rate. That is, each instance is associated with the discounted value of the target reached. If no target is reached, step-level costs are associated with each instance. 
% $x_l=\sum_{\tau=l+1}^T \gamma^{\tau-l} R_{\tau}$

\subsubsection{IBL-TD: A Temporal Difference Mechanism}
IBL-TD is a novel method that involves the integration of the temporal difference (TD) error in RL into the IBL model. The IBL-TD method relies on prediction errors between state valuations, in which an agent iteratively constructs value estimates of states as they are experienced ~\citep{sutton2018reinforcement}, but it also leverages the recognition-based retrieval mechanism of IBLT~\citep{GONZALEZ03}. That is, the IBL-TD model calculates the TD error based on the blended values (Eq. \ref{eq:blended_value}).
The TD error in IBL-TD is formally described by:

\begin{equation}
\begin{array}{l}
    \delta_l = R_l+\gamma \max_A V_{(S_{l+1}, A),t}-V_{(S_l, A_l),t}
\label{eq:IBLTD_err}
\end{array}
\end{equation}

Using this error, the outcome stored in memory for the $l$th instance in an episode is given by the following equation:

\begin{equation}
\begin{array}{l}
    x_{l} \leftarrow V_{(S_l, A_{l}), t} + \alpha \delta_l
    \end{array}
\label{eq:IBLTD_update}
\end{equation}
where $\alpha$ is a step-size parameter.
The full procedural form of the IBL-TD algorithm is specified in Algorithm~\ref{alg:IBLTD}.

\begin{algorithm}[H] 
\caption{IBL-TD Algorithm} 
\label{alg:IBLTD}
\begin{algorithmic}
\STATE \textbf{Initialize}: default utility $x_0\in\mathbb{R}$, global counter $t=1$, step limit $T_{\max} \in \mathbb{N}^+$
% \STATE \quad Initialize default utility $x_0\in\mathbb{R}$ and global counter $t=0$
% \STATE \quad For each $(S,A)\in(\{\mathcal{S}\}\times\{\mathcal{A}\})$ append $(S, A, x_0)$ to $I_{(S,A)}$
%\STATE \quad append $(S, A, x_0)$ to $I_{(S,A)}$ $\forall S \in \mathcal{S}, A \in \mathcal{A}$
% \STATE \quad Initialize global counter $t=0$
\STATE \textbf{Loop for each episode:} 
\STATE \quad Initialize episode step counter $l=1$ and state $S_l$
% \STATE \quad Initialize state $S_l$
\STATE \quad \textbf{Repeat:}
\STATE \quad \quad Choose $A_{l}\in\mathcal{A}$ using $\max_A V_{(S_l,A),t}$ \quad \; (Eq. \ref{eq:blended_value})
\STATE \quad \quad Take action $A_l$, observe $S_{l+1}$ and $R_l$
% \STATE \quad \quad $k\leftarrow (S_l, A_l)$
\STATE \quad \quad $\delta_{l} \leftarrow R_{l}+\gamma \max_A V_{(S_{l+1}, A),t}-V_{(S_l,A_l),t} \,$  (Eq. \ref{eq:IBLTD_err})
% \STATE \quad \quad $\delta_{j} \leftarrow R_{j}+\gamma \cdot \max_A V_{(S_{j+1}, A),t}-V_{(S_j,A),t} \,$ (Eq. \ref{eq:IBLTD_err})
\STATE \quad \quad $x_l \leftarrow V_{(S_l,A_l),t} + \alpha \delta_l \,$ \quad \quad \quad \quad \quad \quad \quad \; \; (Eq. \ref{eq:IBLTD_update})
\STATE \quad \quad Store instance $(S_l,A_l,x_l)$% to $I_k$
\STATE \quad \quad $l \leftarrow l+1$  and $t \leftarrow t+1$
% \STATE \quad \quad $t \leftarrow t+1$
\STATE \quad \quad \textbf{until} $S_l$ is terminal or $l = T_{\max} + 1$
\end{algorithmic}
\end{algorithm}

% %
% \begin{algorithm}[H] 
% \caption{IBL-TD Algorithm} 
% \label{alg:IBLTD}
% \begin{algorithmic}[1]
% \STATE \textbf{Initialize}: default utility $x_0\in\mathbb{R}$, global counter $t=1$, step limit $T_{\max} \in \mathbb{N}^+$
% \STATE \textbf{Loop for each episode:} 
% \STATE \quad Initialize episode step counter $l=1$ and state $s_l$
% \STATE \quad \textbf{Repeat:}
% \STATE \quad \quad Choose $a_{l}\in\mathcal{A}$ using $\max_a V_{(s_l,a),t}$ 
% \STATE \quad \quad Take action $a_l$, observe $s_{l+1}$ and $r_l$
% \STATE \quad \quad $\delta_{l} \leftarrow r_{l}+\gamma \max_a V_{(s_{l+1}, a),t}-V_{(s_l,a_l),t} \,$ 
% \STATE \quad \quad $x_l \leftarrow V_{(s_l,a_l),t} + \alpha \delta_l \,$ 
% \STATE \quad \quad Store instance $(s_l,a_l,x_l)$
% \STATE \quad \quad $l \leftarrow l+1$  and $t \leftarrow t+1$
% \STATE \quad \quad \textbf{until} $s_l$ is terminal or $l = T_{\max} + 1$
% \end{algorithmic}
% \end{algorithm}

%\input{4_exp.tex}
\section{Human Experiments and Evaluation of Models of Credit Assignment} \label{sec:human_exp}
We conducted two behavioral experiments based on interactive browser-based gridworld applications and ran simulations using the IBL and RL models in the same task. The experiments manipulate the level of decision complexity, which is characterized by the relative costs and benefits of the highest value target and the closest distractor to the agent's initial location (i.e., spawn location) in the gridworld ~\cite{NGUYEN2020ICCM}.

\begin{figure}[!htbp]
\centering
 \begin{subfigure}[b]{0.32\linewidth}
        \includegraphics[width=\linewidth]{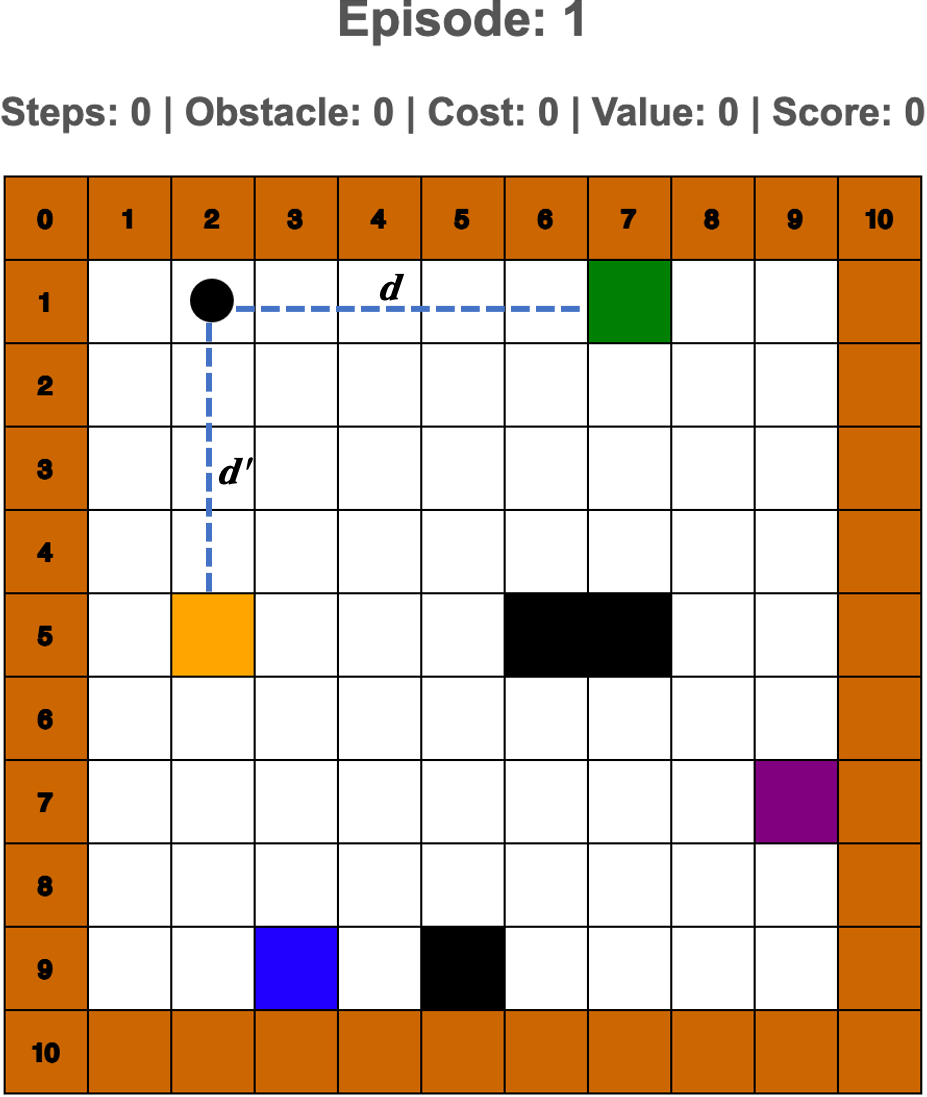}
\caption{\textbf{Simple grid}}
\label{fig:simple}
    \end{subfigure}\hspace{10mm} %or \hspace{0.3\textwidth}
\begin{subfigure}[b]{0.32\linewidth}
        \includegraphics[width=\linewidth]{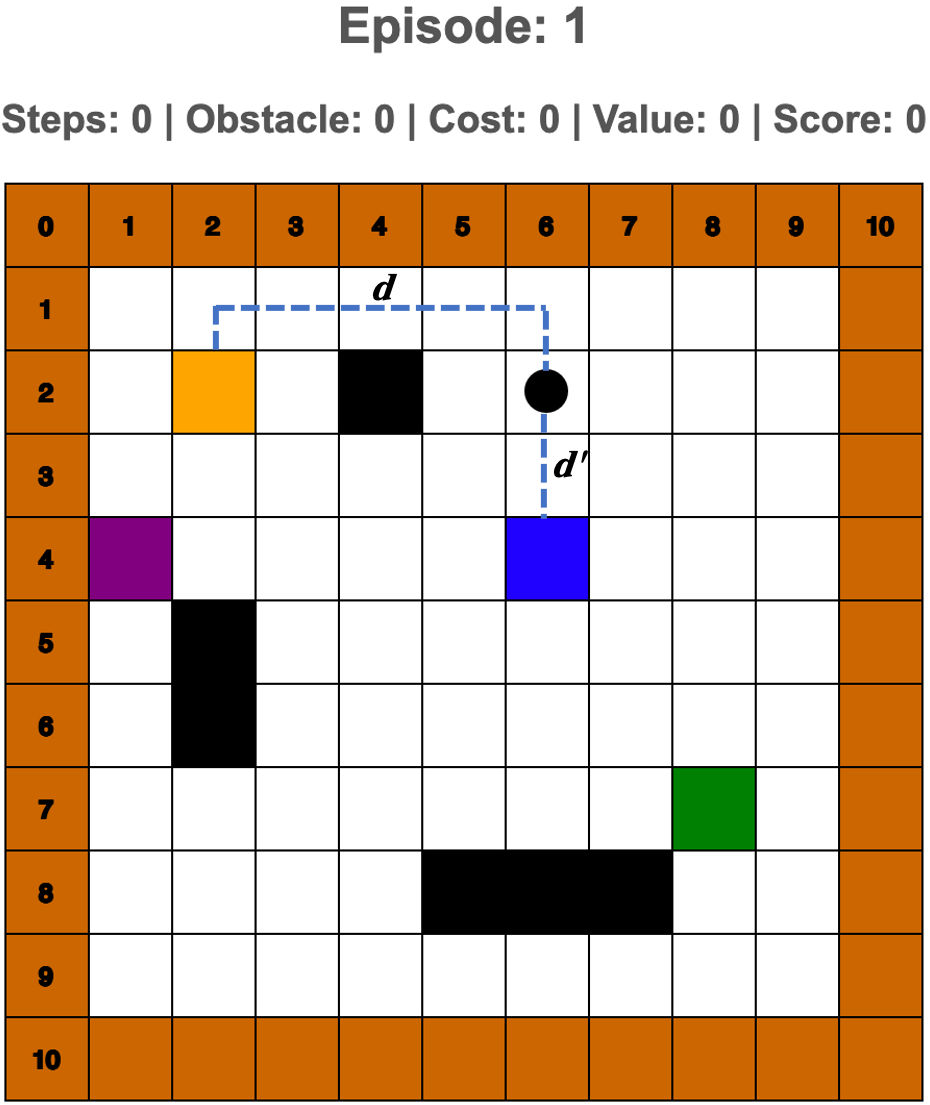}
\caption{\textbf{Complex grid}}
\label{fig:complex}
    \end{subfigure} 
\caption{An example of one of 100 different gridworlds used in simple and complex conditions. In (a), the highest value target is ``green'' and the distractor is ``orange''. In (b), the highest value target is ``orange'', the distractor is ``blue'', and there is a block on the way to the highest-value target.}
\label{fig:conditions}
\end{figure}

Fig.~\ref{fig:conditions} illustrates examples of simple and complex decisions in two grids. Decision complexity was measured precisely by the difference between the distance from an agent's spawn location to the highest value target ($d$) and to the nearest distractor ($d'$), denoted by $\Delta_d = d - d'$. The larger the value of $\Delta_d$, the more complex a decision is: higher values of $\Delta_d$ indicate a tension between consuming the highest reward object with a longer distance $d$ or going for the distractor with the shorter distance $d'$. We considered two levels of decision complexity: ``Simple'' gridworlds with $\Delta_d = 1$ and ``Complex'' gridworlds with $\Delta_d = 4$.

\subsection{Human Experiment Design}
The gridworlds presented to human participants were selected randomly from a set of 100 different grids according to the level of decision complexity in each participant's condition. Participants are required to navigate a grid in each episode with a step limit ($T_{\max}=31$) to reach the targets and maximize their score. The grid remains unchanged across $40$ playing episodes. Their spawn location is fixed in each configuration and determined by the complexity condition. The episode ends when participants first reach one of the four targets or when they reach the 31-step limit without any target consumption. Participants earned points for reaching the target and were penalized for each movement (-0.01) and for walking into a wall or an obstacle (-0.05).
We note that the value was multiplied by 100 to make it easier for participants to interpret their points in the game.

The two experiments used identical gridworlds in the goal-seeking task, providing participants with identical underlying environments, i.e., the target and obstacle locations. 
In Experiment 1, participants were provided with different information regarding the environment compared to Experiment 2, in which they only received limited information. The exact instructions are found in the Supplementary materials.
% Full details on the methods can be found in the Methods section ~\ref{sec:methods} and exact instructions are found in the Supplementary materials.

\subsubsection{Experiment 1: Grid Information.}
\begin{figure}[!htbp]
\centering
\includegraphics[width=0.3\linewidth]{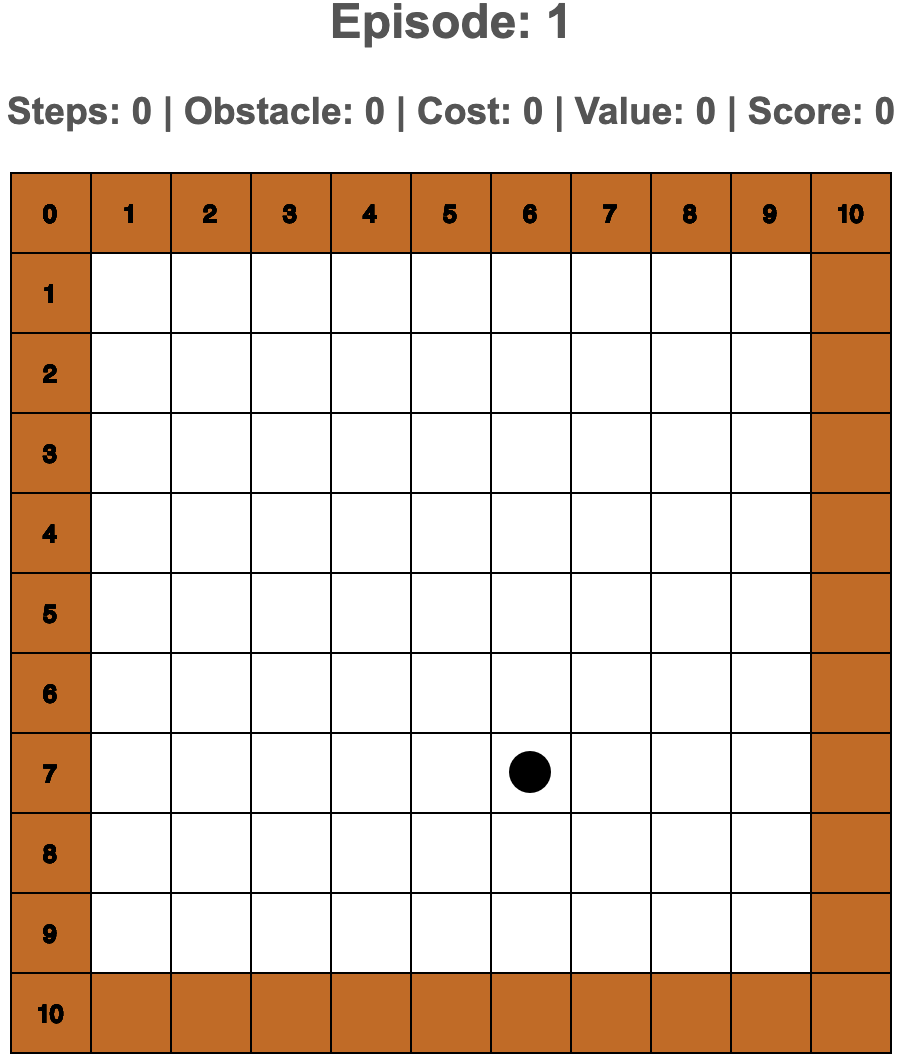}
\caption{Illustration of the goal-seeking task in the gridworld.}
\label{fig:grid_task}
\end{figure}

In Experiment 1, participants viewed an interface as shown in Fig.~\ref{fig:grid_task}, and navigated through the grid by making a sequence of decisions (i.e., move up, down, left, or right) to locate the target with the highest value\footnote{Demo of the game in Experiment 1:~\url{http://janus.hss.cmu.edu:3001/}}. Participants observed their current position within the grid (i.e., a black dot). After making a move, the content of the new location (i.e., whether it was an empty cell, an obstacle, or a target) was revealed. 

We have 100 gridworlds at the same set of levels. In the end, 64 (out of 100) Simple and 62 (out of 100) Complex configurations were randomly assigned to the participants.
The study is preregistered with Open Science Framework\footnote{Pre-registration link: \url{https://osf.io/s4prw/?view\_only=16d8e425d9644f158c82ed32a5b2a35e}{}. According to the preregistration, we also collected human data in the third condition called ``Random'', which is essentially a mixed condition of ``Simple'' and ``Complex'' gridworlds. We did not observe any results standing out in this condition, and thus, we did not report them in this work.}.
% We manipulated the levels of decision complexity: ``Simple'' ($\Delta_d = 1$), ``Complex'' ($\Delta_d = 4$), and ``Random'' ($\Delta_d$ varies from 1 to 4).

\textit{Procedure.} Before starting the task, participants were given a practice episode in a smaller grid (size $6\times6$). During the main game, participants were asked to complete 40 episodes of the same gridworld environment. Finally, participants were asked to recall the targets' position and their judgment about the highest value target. 
The study took 15-30 minutes to complete.
Participants were compensated a base payment of \$1.50  and earned up to \$3.00 in a bonus payment according to their accumulated score. 

\textit{Participants.} A total of 305 participants were recruited from Amazon Mechanical Turk and completed the study. Of those 102 were assigned to the ``Simple'' condition (mean $\pm$ standard deviation age: 36.5 $\pm$ 10.3; 34 female) and 104 performed the ``Complex'' condition (37.9 $\pm$ 10.8, 37 female).
% and 99 played the ``Random'' condition (36.6 $\pm$ 9.6, 38 female). 

\subsubsection{Experiment 2: Restricted Grid Information.}
\begin{figure}[!htbp]
\centering
\includegraphics[width=0.7\linewidth]{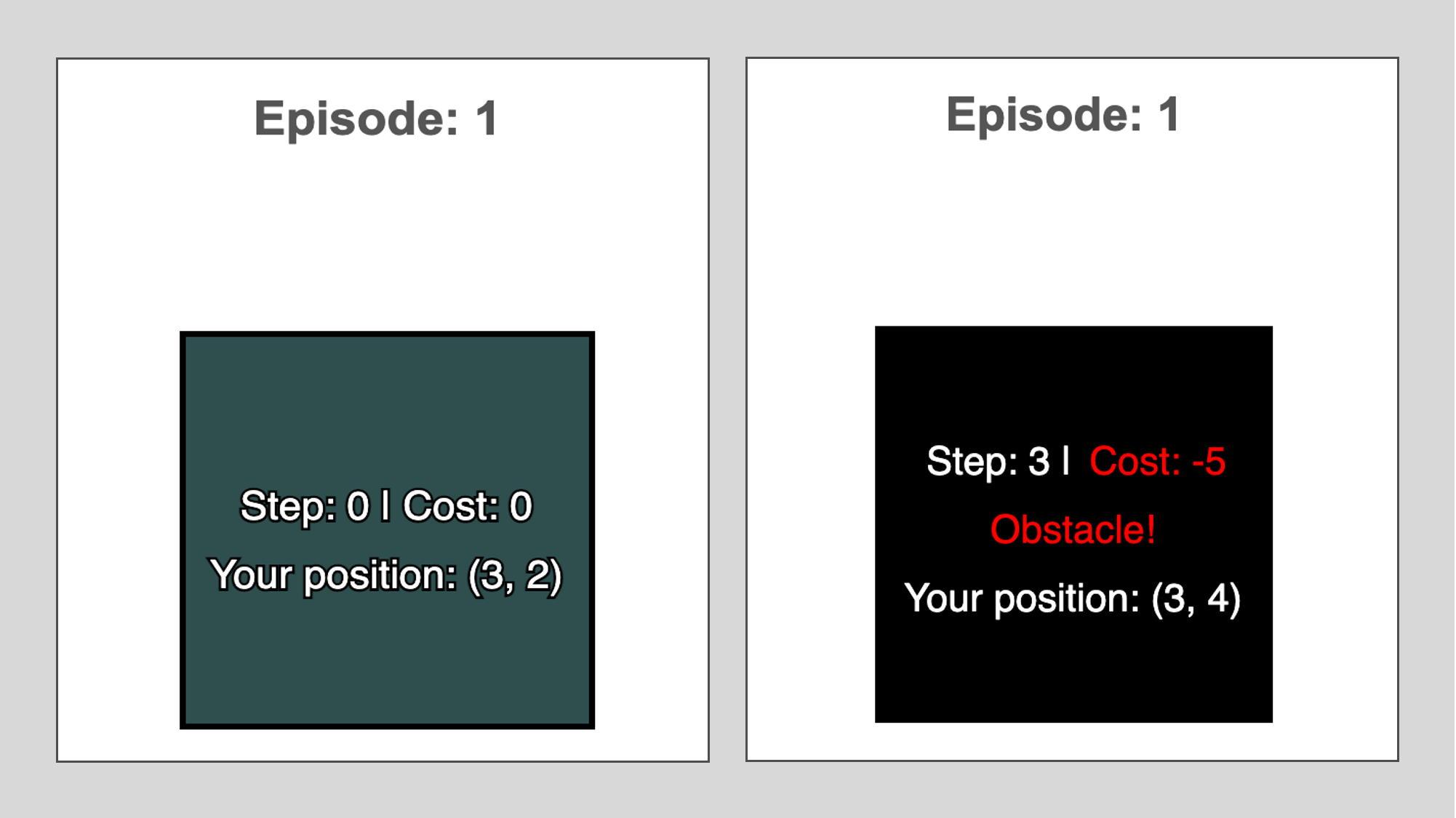}
\caption{Illustration of the goal-seeking task in the restricted gridworld information when the game is started (left), and when an obstacle is encountered (right).}
\label{fig:onestep_task}
\end{figure}

In Experiment 2, participants were provided with limited information, as shown in Fig.~\ref{fig:onestep_task}. Participants were presented with only one cell at each step, which informs them about their current ($x$, $y$) position, the count of steps taken, and the immediate cost or reward of the previous step\footnote{Demo of the game in Experiment 2: \url{http://janus.hss.cmu.edu:3006/}}. All information of obstacles, the value of the targets, as well as the shape and size of the grid were concealed. Similar to Experiment 1, participants started at an initial position determined by the decision complexity, and were tasked with navigating the gridworld to find the highest value target. The costs and rewards were identical to those in Experiment 1. Other than the information difference, the task and gridworlds used were identical to those of Experiment 1.

We considered the same gridworld design and configuration of the decision complexity as in Experiment 1.
The study is preregistered with Open Science Framework\footnote{Pre-registration link: \url{https://osf.io/m7quh/?view\_only=2728f5ccde1f41cdabf256393f3bd6a2}. Similar to Experiment 1, we did not report the results in the ``Random'' condition in this work.}.

\textit{Procedure.} As opposed to Experiment 1, participants did not have a practice session. Instead, after the instructions, they were asked to start the main game, which entails 40 episodes of the same gridworld environment. Finally, participants were asked about their memory of the targets' position as well as their judgment about the highest value target. 
The study took 15-30 minutes to complete. The same base and bonus payment were applied to this experiment.

\textit{Participants.} A total of 301 participants were recruited from Amazon Mechanical Turk and completed the study. Of those 99 were assigned to the ``Simple'' condition (age: 37.7 $\pm$ 11.8; 40 female) and 95 performed the ``Complex'' condition (38.2 $\pm$ 11.3, 30 female).
% and 107 played the ``Random'' condition (36.9 $\pm$ 11, 44 female).

\subsection{Model Simulation}
For comparison, we considered Q-learning \cite{watkins1992q,sutton2018reinforcement}, a foundational TD-based algorithm. 
The Q-learning algorithm maintains estimates of optimal state-action pairs using the $Q(S,A)$ value that represents the expected return of taking action $A$ in state $S$. At each step, the agent selects the action ($\epsilon$-greedy) that maximizes the expected return, e.g. $\max_A Q(S,A)$. 
For estimate updates, the TD error is given by $\delta_l=R_l+\gamma \max_A Q(S_{l+1}, A)-Q(S_l,A_l)$ and the update at each step is $Q(S_{l}, A_l)\xleftarrow{}Q(S_{l}, A_l)+\alpha\delta_l$.

\paragraph{Parameter Selection}
We ran simulations on the IBL-Equal, IBL-Exponential, IBL-TD, and Q-learning models to collect synthetic data for each model using its default parameters. The rationale for using the default parameters rather than fitting the parameters to the data is that such results can represent pure predictions from each model reflecting their true nature. 

It is worth noting that in the case of IBL-TD and Q-learning, they have no parameters prescribed for the temporal difference mechanism. Indeed, the method was developed for \emph{reward maximizing} agents and thus encourages a search over the parameter space for use. As such, the default parameters for these models are derived from performing a sweep over the parameter space, with the best parameter configuration being used as the default. In particular, this was accomplished via Tune \cite{liaw2018tune}, where we selected the best model after 1,000 experimental trials. We select the set of parameters that maximizes the proportion of episodes wherein the agent reaches the highest value target. Moreover, all models are instantiated with default utilities or state-action valuations of 0.4. Such a default was found to increase performance for all models when compared with a default of 0.0, as it encourages environment exploration.
The full set of our parameter selections is described in Table \ref{tab:params}.

\begin{table}[!htbp]
\centering
\resizebox{.6\textwidth}{!}{% 
\begin{tabular}{ll|ccccc}
\toprule
         & Model &  $\sigma$ &  $d$ &  $\gamma$ &  $\alpha$ &  $\epsilon$   \\
\midrule
    \multirow{4}{*}{Simple}
       & IBL-Equal &  0.25 &  0.50 &     --- &     --- &     ---   \\
 & IBL-Exponential &    0.250 &  0.50 &  0.990 &     --- &     ---   \\
           &IBL-TD &   0.049 &  0.95 &  0.986 &  0.824 &     ---  \\
       & Q-learning &     --- &     --- &  0.997 &  0.839 &   0.002   \\

\midrule
    \multirow{4}{*}{Complex}
    & IBL-Equal     &  0.25 &  0.5 &     --- &     --- &     ---      \\
  & IBL-Exponential &  0.25 &    0.5 &  0.990 &     --- &     ---   \\
           & IBL-TD &    0.038 &  0.886 &  0.999 &  0.838 &     ---   \\
       & Q-learning &     --- &     --- &  0.977 &  0.865 &  0.022   \\
% \midrule
\bottomrule
\end{tabular}
}
    \caption{Parameters used for each model in the default settings. IBL-TD and Q-learning parameters are fit to maximize accuracy, IBL-Equal and IBL-Exponential utilize the default prescribed in the IBLT literature \protect\cite{GONZALEZ03}, with a standard default discount rate for IBL-Exponential ($\gamma=0.99$).}
    \label{tab:params}
\end{table}

\paragraph{Procedure.} A total of 378 independent runs (64 Simple and 62 Complex configurations with three runs on each) for each of the models described above were run to perform the same gridworld task as human participants. For each level of complexity, we instantiate the models in the corresponding gridworld configurations from the human experiments: 62 complex and 64 simple gridworld configurations. For each level of complexity and for each of the gridworld configurations, three runs of each model type were executed from scratch for 40 episodes. Equivalent to the human experiment, the agent models were allotted 31 steps to reach a target. We calculated the same evaluation metrics that are described in the next section, using the data of 40 episodes from 300 agents per level of complexity.

\subsection{Evaluation Metrics}
In both experiments, we compared the performance of each model to that of humans in terms of two metrics: (i) Proportion of Maximization (i.e., \textit{PMax}): the proportion of episodes wherein the individual (or the agent) obtains the highest value target, and (ii) Proportion of Optimal choice (i.e., \textit{POptimal}): the proportion of episodes wherein the individual (or the agent) not only reaches the highest value target but also does so by taking the minimal number of steps possible (i.e., via a shortest path).
We also measured the difference between the average performance of each model and the human data with respect to the corresponding metrics for each complexity condition.
\section{Experimental Results}
Here we describe the obtained results through human study as well as simulation, using the various models described above, to assess which credit assignment mechanism is most closely aligned with human behavior and which can better achieve optimal performance.  

\subsection{Experiment 1: Grid Information}
Table \ref{tab:human_no_fit} presents the average PMax, POptimal and difference of the averages for each model compared to human participants (i.e., Model - Human).

Overall, we see that in both simple and complex environments, IBL-Equal is the best performer in terms of PMax and IBL-TD is the best in terms of POptimal. Regarding the similarity of Pmax to humans, the TD models are worse than humans in simple environments, but all models are better than humans in complex environments. In terms of POptimal, all models perform worse than humans in simple environments and in complex environments, except for IBL-TD that on average performs slightly better than humans. However, clearly, the average behavior is highly uninformative regarding the dynamics of learning during the 40 episodes. We, therefore, turn our attention to learning curves from hereon.

\begin{table}[!htbp]
\centering
\resizebox{.99\textwidth}{!}{% 
\begin{tabular}{@{}lrrrrlrrrr@{}}
\toprule
 Decision Complexity & \multicolumn{4}{c}{Simple} &  & \multicolumn{4}{c}{Complex} \\
 \cmidrule(l){2-10} 
 & \begin{tabular}[c]{@{}r@{}}PMax \\ Avg.\end{tabular} & \begin{tabular}[c]{@{}r@{}}PMax\\ Diff.\end{tabular} & \begin{tabular}[c]{@{}r@{}}POpt.\\ Avg.\end{tabular} & \begin{tabular}[c]{@{}r@{}}POpt.\\ Diff.\end{tabular} & \multicolumn{1}{r}{} & \begin{tabular}[c]{@{}r@{}}PMax\\ Avg.\end{tabular} & \begin{tabular}[c]{@{}r@{}}PMax\\ Diff.\end{tabular} & \begin{tabular}[c]{@{}r@{}}POpt.\\ Avg.\end{tabular} & \begin{tabular}[c]{@{}r@{}}POpt.\\ Diff.\end{tabular} \\ \midrule
Human & 0.71 & --- & 0.66 & --- &  & 0.48 & --- & 0.43 & --- \\ \cdashlinelr{1-10}
      IBL-Equal &           \textbf{0.80} &                  0.09 &              0.14 &                     -0.52 & &          \textbf{0.73} &                  0.25 &              0.37 &                     -0.06 \\
IBL-Exponential &           0.79 &                  0.08 &              0.18 &                     -0.48 &  &         0.67 &                  0.19 &              0.42 &                     -0.01 \\
         IBL-TD &           0.68 &                  -0.04 &              \textbf{0.49} &                     -0.17 & &          0.62 &                  0.14 &              \textbf{0.44} &                     0.01 \\
     Q-learning &           0.67 &                  -0.05 &              0.46 &                     -0.20 &  &         0.61 &                  0.13 &              0.40 &                     -0.03 \\ \bottomrule
\end{tabular}

% \begin{tabular}{@{}llrrrrlrrrr@{}}
% \toprule
%  &  & \multicolumn{4}{c}{Simple} &  & \multicolumn{4}{c}{Complex} \\ \cmidrule(l){3-11} 
%  &  & \begin{tabular}[c]{@{}r@{}}PMax \\ Avg.\end{tabular} & \begin{tabular}[c]{@{}r@{}}RMSE\\ Err.\end{tabular} & \begin{tabular}[c]{@{}r@{}}POpt.\\ Avg.\end{tabular} & \begin{tabular}[c]{@{}r@{}}RMSE\\ Err.\end{tabular} & \multicolumn{1}{r}{} & \begin{tabular}[c]{@{}r@{}}PMax.\\ Avg.\end{tabular} & \begin{tabular}[c]{@{}r@{}}RMSE\\ Err.\end{tabular} & \begin{tabular}[c]{@{}r@{}}POpt.\\ Avg.\end{tabular} & \begin{tabular}[c]{@{}r@{}}RMSE\\ Err.\end{tabular} \\ \midrule
% \multicolumn{1}{c}{} & Human & 0.71 & --- & 0.66 & --- &  & 0.48 & --- & 0.43 & --- \\
% %  &  & \multicolumn{1}{l}{} & \multicolumn{1}{l}{} & \multicolumn{1}{l}{} & \multicolumn{1}{l}{} &  & \multicolumn{1}{l}{} & \multicolumn{1}{l}{} & \multicolumn{1}{l}{} & \multicolumn{1}{l}{} \\
% \parbox[t]{2mm}{\multirow{4}{*}{\rotatebox[origin=c]{90}{Model}}}
% & IBL-Equal & 0.80 & 0.11 & 0.14 & 0.53 &  & 0.73 & 0.27 & 0.37 & 0.07 \\
%  & IBL-Exponential & 0.79 & 0.12 & 0.18 & 0.50 &  & 0.67 & 0.21 & 0.42 & 0.05 \\
%  & IBL-TD & 0.68 & 0.17 & 0.49 & 0.29 &  & 0.62 & 0.22 & 0.44 & 0.24 \\
%  & Q-learning & 0.67 & 0.17 & 0.46 & 0.30 &  & 0.61 & 0.21 & 0.40 & 0.19 \\ \bottomrule
% \end{tabular}
}
\caption{\textbf{Experiment 1:} Average performance and difference (Model - Human) for each model compared to the human data in each condition.
The boldface indicates the best results among the models in terms of average PMax and POptimal. In the simple conditions, the average PMax and POptimal of human data are 0.71 and 0.66, respectively, while in the complex settings, the humans' average PMax and POptimal are 0.48 and 0.43, accordingly. }
\label{tab:human_no_fit}
\end{table}

\begin{figure}
    \centering
    \includegraphics[width=\textwidth]{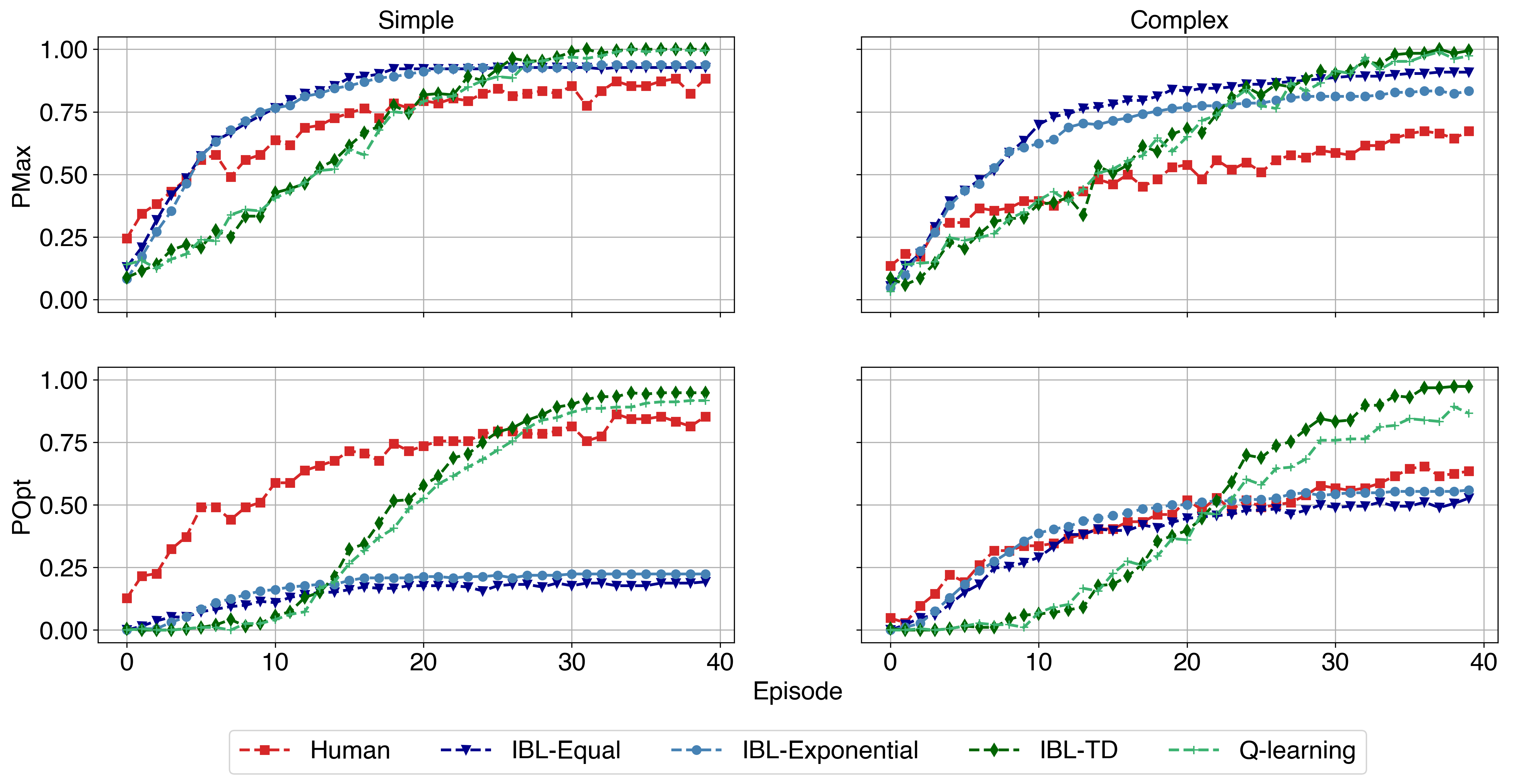}
    \caption{\textbf{Experiment 1 results:} PMax and POptimal with the default parameters.}
    \label{fig:default_parameter_learning_curve}
\end{figure}

We plotted the PMax and POptimal averaged over each episode. The results are shown in Fig.~\ref{fig:default_parameter_learning_curve}. The figure depicts more clearly the similarities and gaps between average model performance over episodes compared to that of human participants. By the observation of these learning curves, it is immediately obvious that humans (red lines) are affected by decision complexity, while the effect of complexity on the models is minimal. In particular, the TD models (green lines) appear to be unaffected by decision complexity (all statistical results are presented in the Supplementary Information).

These figures also expose significant differences between the way models and humans learn. IBL-Exponential and IBL-Equal (blue lines) capture the shape of the learning curves for PMax in simple decisions quite well, but they fail to reach comparable levels of POptimal in simple decisions. In complex decisions, these models outperform the PMax of humans but have a closer fit to humans' POptimal.

We further notice that the POptimal of IBL-Equal and IBL-Exponential agents is higher in the complex than in the simple condition, which is the reverse pattern observed in humans' POptimal. 
We speculate that such enhancement of the models in complex settings is due to the environment design. That is, in the complex settings, there is an obstacle located on the direct path towards the highest value target as exemplified in Fig~\ref{fig:complex}. Given that the agents are penalized when hitting an obstacle and such negative consequence is reinforced in the IBL agents' memory, they might learn a better way to find the highest value target by avoiding the obstacle. These results are further elaborated in Section~\ref{sec:analysis}.  

The learning paths of IBL-TD and Q-learning (green lines) are particularly interesting. These models start generally with very poor behavior compared to other models and humans. However, these models show a sharp learning curve and quickly ``catch up'' to human levels of performance in simple condition, and they even surpass the level of human performance in complex decision environments. It is well-known that TD  approaches learn by bootstrapping, that is, they can perform updates based on current estimates before the final outcome is revealed. Hence, the TD models require a large amount of training to reach the same level of human performance in games \citep{tsividis2021human}. By contrast, the Equal and Exponential mechanisms adjust the expected value of all decisions made within each episode after the final outcome of the episode is known, which can be attributed to matching the initial performance of humans in the task. 

In recognizing the differences between the information provided to humans in the experiment and the information given to the models, we observe a practical asymmetry. Humans have access to more information than models do in solving this task. Specifically, participants in our experiment knew about the shape and size of the grid environment, they were informed of the number of targets, they were able to perceive their own location within the boundaries of the grid environment, and they were able to visualize the obstacles after each move and the walls surrounding the environment (Fig.~\ref{fig:grid_task}). Such informational advantages given to human participants in the experiment could explain a better initial performance in the task compared to the TD models in particular. Experiment 2 aims at addressing this asymmetry of information.

\subsection{Experiment 2: Restricted Grid Information}
We recall that to address the asymmetry of information, in Experiment 2 we aimed to provide participants with the same information as it is available to the models. That is, participants were presented with only one cell at a time, which informed them about their current (x, y) position, the immediate cost or reward of the previous step, and the count of steps already taken.

Table \ref{tab:human_no_fit_exp2} reports the average PMax, POptimal, and the difference between the models and human data (all statistical results are presented in the Supplementary Information). It is important to note that the model data regarding the average PMax and POptimal reported in Experiment 2 are identical to those reported in Experiment 1. Simulations were not run again because the models are the same and the only difference is in the human data. Thus, in both simple and complex environments, IBL-Equal is the best performer in terms of Pmax, and IBL-TD is the best performer in terms of POptimal.

\begin{table}[!htpb]
\centering
\resizebox{.99\textwidth}{!}{% 
\begin{tabular}{@{}lrrrrlrrrr@{}}
\toprule
Decision Complexity & \multicolumn{4}{c}{Simple} &  & \multicolumn{4}{c}{Complex} \\ \cmidrule(l){2-10} 
 & \begin{tabular}[c]{@{}r@{}}PMax \\ Avg.\end{tabular} & \begin{tabular}[c]{@{}r@{}}PMax\\ Diff.\end{tabular} & \begin{tabular}[c]{@{}r@{}}POpt.\\ Avg.\end{tabular} & \begin{tabular}[c]{@{}r@{}}POpt.\\ Diff.\end{tabular} & \multicolumn{1}{r}{} & \begin{tabular}[c]{@{}r@{}}PMax \\ Avg.\end{tabular} & \begin{tabular}[c]{@{}r@{}}PMax.\\ Diff.\end{tabular} & \begin{tabular}[c]{@{}r@{}}POpt.\\ Avg.\end{tabular} & \begin{tabular}[c]{@{}r@{}}POpt.\\ Diff.\end{tabular} \\ \midrule
Human & 0.63 & --- & 0.55 & --- &  & 0.24 & --- & 0.17 & --- \\ \cdashlinelr{1-10}
      IBL-Equal &           \textbf{0.80} &                  0.17 &              0.14 &                     -0.41 &&           \textbf{0.73} &                  0.49 &              0.37 &                     0.20 \\
IBL-Exponential &           0.79 &                  0.17 &              0.18 &                     -0.37 &&           0.67 &                  0.43 &              0.42 &                     0.25 \\
         IBL-TD &           0.68 &                  0.05 &              \textbf{0.49} &                     -0.06 &&           0.62 &                  0.38 &              \textbf{0.44} &                     0.27 \\
     Q-learning &           0.67 &                  0.04 &              0.46 &                     -0.09 &&           0.61 &                  0.38 &              0.40 &                     0.23 \\ \bottomrule
\end{tabular}
}
\caption{\textbf{Experiment 2:} Average performance and difference (Model - Human) for each model compared to the human data. Bold face indicates the best results among the models in terms of average PMax and POptimal. The average PMax and POptimal of human data in the simple conditions are 0.63 and 0.55 respectively, whereas in the complex settings, the average PMax and POptimal are 0.24 and 0.17, accordingly.}
\label{tab:human_no_fit_exp2}
\end{table}

Relative to Experiment 1, it is clear that restricting information for humans makes their task more difficult. Humans reached the maximum target less often than in Experiment 1, when confronted with simple decisions (PMax of 0.71 in Experiment 1 vs. 0.63 in Experiment 2) and with complex decisions (PMax of 0.48 in Experiment 1 vs. 0.24 in Experiment 2). Similarly, humans followed the optimal path less often when restricted to less information, compared to Experiment 1 in simple decisions (POptimal= 0.66 in Experiment 1 vs. 0.55 in Experiment 2), and particularly in complex ones (POptimal of 0.43 in Experiment 1 vs. 0.17 in Experiment 2). 

Given the difference in human performance, the models' performance in relation to the human data changed. In the context of Experiment 2, the models are all better than the human participants in terms of PMax, in making simple, but particularly in making complex decisions. In terms of POptimal, the models are all worse than humans in simple environments, but they are exceedingly better than humans in complex decisions.

\begin{figure}[!htpb]
    \centering
    \includegraphics[width=\textwidth]{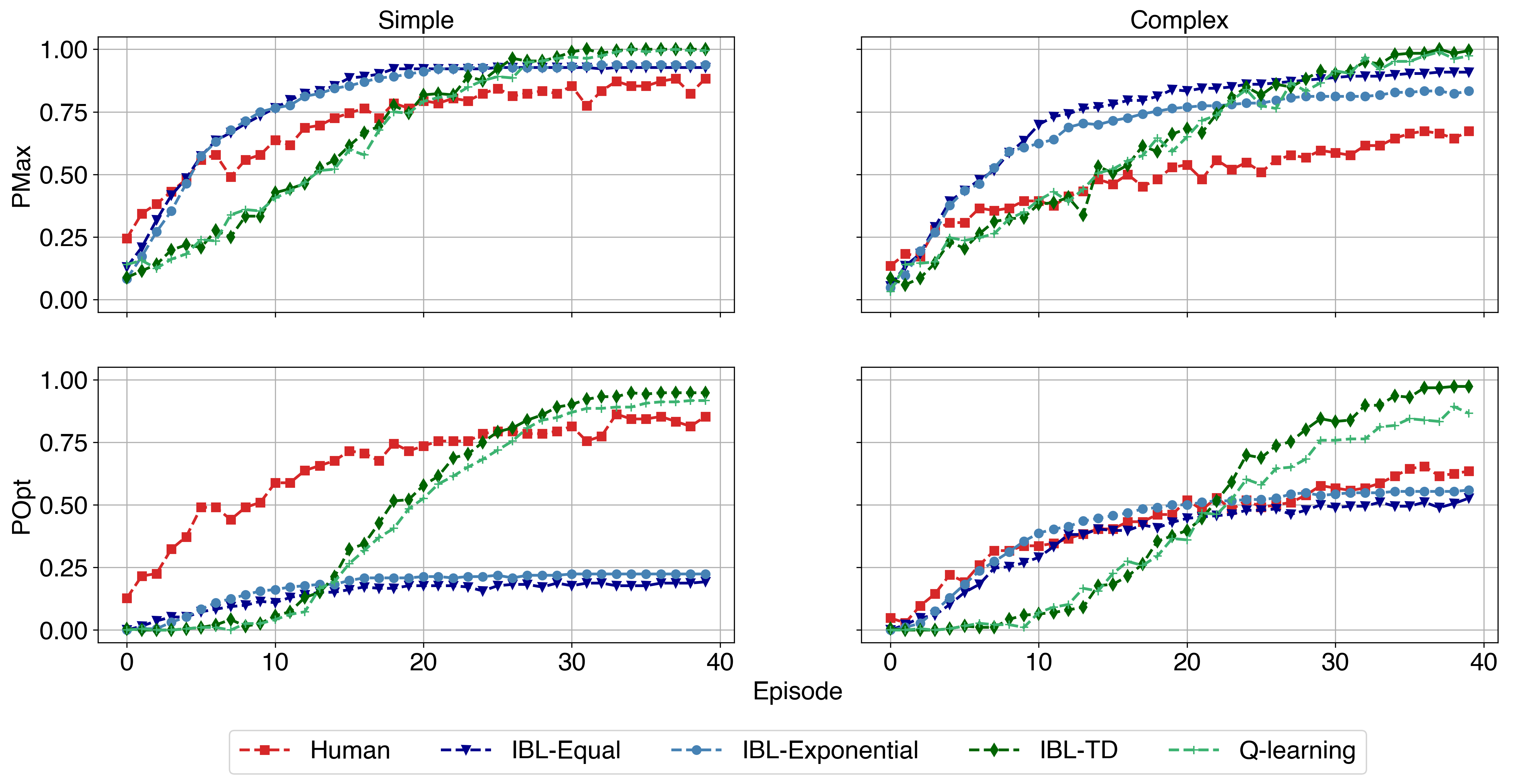}
    \caption{\textbf{Experiment 2 results:} PMax and POptimal with the default parameters. Model results are identical to those in Figure \ref{fig:default_parameter_learning_curve}.}
    \label{fig:default_parameter_exp2_learning_curve}
\end{figure}

In the learning curves presented in Fig.~\ref{fig:default_parameter_exp2_learning_curve} it is possible to observe again, the significant impacts of the complexity on human performance. The PMax figures (top panels) illustrate how the models no longer compare to humans' PMax as in Experiment 1, particularly in the complex condition. Human performance deteriorated significantly in complex compared to simple grids when there is restricted information.

Importantly, humans did not show an initially slow POptimal as predicted by the IBL-TD and Q-learning models. The initial large gap between human POptimal and IBL-TD and Q-learning models persists, despite the symmetry of information in the models and humans. Furthermore, given a lower human POptimal in complex grids, the final gap between the IBL-TD and Q-learning and human POptimal is significantly larger in this experiment than in Experiment 1. That is, these models outperform humans in the final episodes and are robust to decision complexity.

\subsection{Summary of results so far}
The analyses in the sections above elucidate the major differences between the predictions of models with three credit assignment mechanisms and human performance. Importantly, the differences are not due to the asymmetry of information provided to humans compared to the information available to the models. Comparing the results of Experiment 2, where the information available to humans was symmetrical to the information given to the models, we observed additional challenges for the models: models are unable to predict the complexity effects that humans confront under restricted information.  

A pattern that seems consistent across the two experiments is that, on average, IBL-Equal obtains the highest PMax regardless of the decision complexity, and IBL-TD achieves the highest POptimal on average. Neither IBL-Equal nor IBL-Exponential are able to reach the POptimal that humans achieve. IBL-TD and Q-learning reach and outperform the level of Poptimal in humans, particularly in complex settings. However, the TD models are unable to capture the initially high POptimal level of humans. Specifically, in complex settings wherein humans' PMax and POptimal are much lower than in simple settings, the TD credit assignment method surpasses the POptimal of humans. This is accentuated in situations in which humans have limited information while performing the task (Experiment 2).

% %Connect to the human fitting section
As a reminder, the results presented above are pure predictions from the models. The models' parameters were not fit to the human data and human data was not used in any way to inform the models. Thus, the observed patterns of models' results are expected to improve when the models are calibrated to human data. To that end, we performed a fitting exercise for each of the models on the average PMax. We ran analogous simulation experiments after having determined the parameters of the models that best-fitted humans' average PMax values. The results presented in the Supplementary material, Section~\ref{sec:human_fit}, showed that indeed the models are able to capture the PMax more closely after the model parameters are fit to human data; however, some of the challenges regarding the initial underperformance of the TD models' compared to human data remained even after finding the best fitting parameters.

\section{Challenges and Opportunities for Human-like AI Agents: A behavioral Analysis} \label{sec:analysis}
The results above expose the limits and gaps of temporal credit assignment mechanisms in two well-known approaches: cognitive IBL models and the Q-learning algorithm for modeling human decisions in a goal-seeking navigation task with feedback delays. While the IBL-Equal and IBL-Exponential models emulate the learning curves of the PMax of participants, we observe a particularity of the learning curves in the TD-based models. These models produce consistent underperformance compared to humans in the initial episodes. At the same time, the TD-based models are better than the IBL-Equal and IBL-Exponential at capturing the POptimal of human participants, and the TD models often outperform humans in the final episodes. 

In this section, we conduct in-depth analyses of humans' and models' behavior in both experiments with the intention of informing the development of human-like AI agents and the design of systems that can help improve human decisions and learning. We compare the data obtained from human players to that of the models in terms of various \textit{process metrics} to shed light on the aspects of human behavior that these models fail to capture along with the characteristics that hinder humans' optimal performance.

% %------------
\subsection{Challenges for Building Human-like AI Agents}
% The investigations in the sections above exposed that humans are sensitive to the complexity of the task, particularly with restricted information (Experiment 2).
This section presents different reasons why the models are unable to capture the initial learning process of humans. We specifically examined different aspects concerning behavior and decision making processes.

\subsubsection{Target Consumption by Decision Complexity} \label{subsec:target_consumption}
\begin{figure}[!htbp]
\centering
\includegraphics[width=0.99\linewidth]{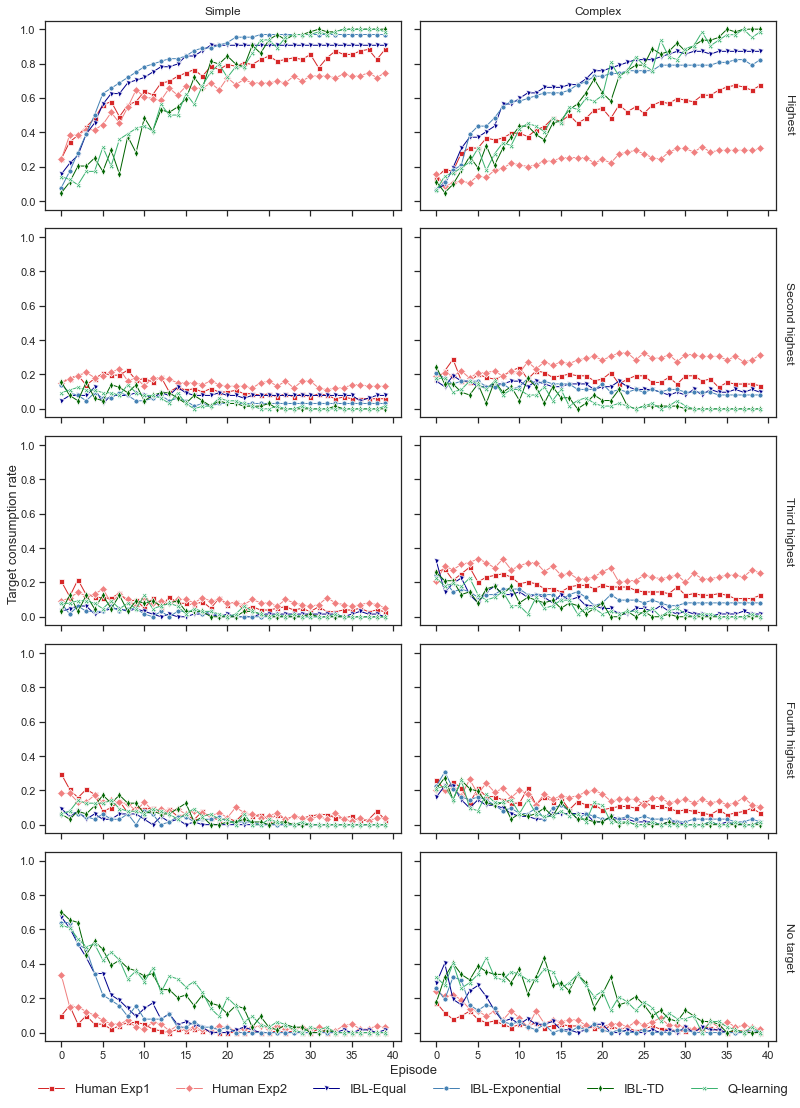}
\caption{The target consumption distribution of humans and agents over time in the simple and complex conditions.}
\label{fig:target_distribution}
\end{figure}

Fig.~\ref{fig:target_distribution} provides the proportion of consuming each of the four targets by their value and the proportion of no target consumption within each episode over the 40 episodes for simple and complex decisions, including humans and models in both experiments. Comprehensive statistical analyses are summarized in the supplementary material (ANOVA Table~\ref{tab:anova_target_consumption} and \ref{tab:tukey_consumption}).

Observations from the figure and statistical results show a significant effect of decision complexity on the proportion of times that the highest value target is consumed by humans and models. Humans and models are able to reach the highest value targets more often in simple than complex decision environments, but the effect of decision complexity on humans is more significant. 
 
Moreover, posthoc analyses with Tukey HSD corrections confirm that the mean rate of finding the highest value target of humans playing the simple ($M=0.63, SD=0.48$) and complex settings ($M=0.24, SD=0.42$) in Experiment 2 was significantly lower than that of humans playing the simple ($M=0.71, SD=0.45$) and complex ($M=0.48, SD=0.49$) conditions in Experiment 1. The results signify that in addition to being susceptible to decision complexity, humans are noticeably susceptible to the degree of information provided in the goal-seeking navigation task (in accordance with findings in prior research~\citep{marusich2016effects}). A similar effect of complexity is observed in almost all lower value targets (i.e., second, third, and fourth highest, see Supplementary ANOVA Table~\ref{tab:anova_target_consumption} for more details).

We additionally observed a distinction in the proportion of times no target was found.  The rates of failure to find a target by the models are significantly higher (especially in the earlier episodes), than those of humans. Furthermore, the proportion of times the TD models were not able to find any target is not influenced by decision complexity, whereas the proportion of times that humans are unable to find a target in an episode is significantly affected by the decision complexity in Experiment 1 and Experiment 2, as reported in supplementary material, ANOVA Table~\ref{tab:anova_target_consumption}.

% %---------------
\subsubsection{Redundancy and Immediate Redundancy} \label{subsec:redundancy}
To understand why the TD models explored the initial episodes without finding any targets, we examined the redundancy of movements made by the agents and humans while navigating the environment. Redundancy of one’s trajectory is defined by the ratio of the number of revisited locations to the number of uniquely visited locations. We also measured immediate redundancy, which is the extent to which agents decide to turn back to the place where they came from. 

\begin{figure}[!htbp]
\centering
\includegraphics[width=0.99\linewidth]{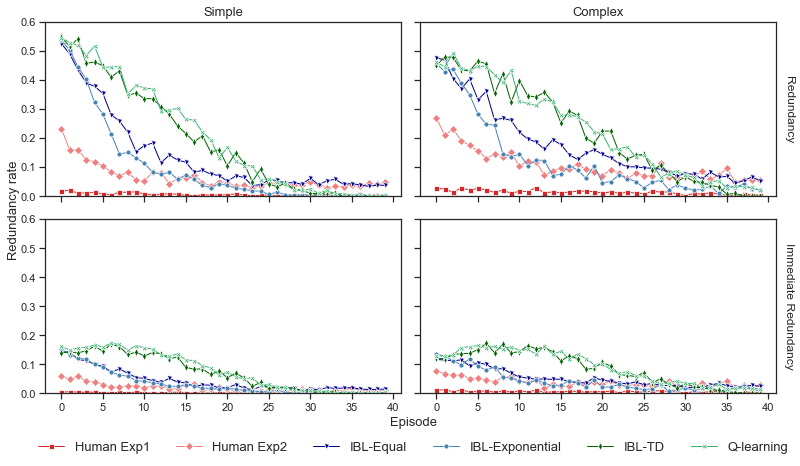}
\caption{Proportion humans and agents that made redundancy and immediately redundant steps over time in each condition.}
\label{fig:redundancy}
\end{figure}

Fig.~\ref{fig:redundancy} shows the proportion of redundant and immediately redundant visits made by humans and models by the level of decision complexity. 
Regarding human redundant actions, we observe that humans in Experiment 1 made almost zero redundant visits from beginning to end, while they made significantly more redundant visits in Experiment 2, when information was limited. The statistical tests further support this observation by indicating that there was no significant effect of the decision complexity on the redundancy rate of humans playing the task in Experiment 1, but with restricted information in Experiment 2, humans made significantly more redundant visits in complex gridworlds than in simple ones (see Supplementary ANOVA Table~\ref{tab:anova_redundancy}).

Post-hoc comparisons using Tukey's test specified that the mean redundancy rate made by humans with restricted information in the simple ($M=0.06, SD=0.15$) and complex settings ($M=0.10, SD=0.19$) of Experiment 2 was significantly higher than those playing the simple ($M=0.006, SD=0.03$) and complex settings ($M=0.01, SD=0.06$) of Experiment 1 (see ANOVA Table \ref{tab:tukey_redundancy}). 

These observations also hold true for the immediate redundant actions. The proportion of immediate visits by humans was not significantly influenced by the decision complexity in Experiment 1. Conversely, in the limited information context of Experiment 2, we find a significant effect of decision complexity on the proportion of times they make immediately redundant movements (see Supplementary Table ANOVA~\ref{tab:anova_redundancy}).

The results show higher proportions of redundant and immediate redundant decisions made by the models compared to humans, particularly in the first episodes (Supplementary Table~\ref{tab:redundancy_stat} and Fig.~\ref{fig:redundancy_2} provide descriptive statistics and an average view of the level of redundancy along with immediate redundancy in the initial and later episodes). From Fig.~\ref{fig:redundancy}, it is clear that the TD models have the highest redundancy and immediate redundancy rates compared to all the models in approximately the first half of the playing episodes in both levels of complexity. 
Furthermore, the rates of taking redundant and immediately redundant visits by IBL-TD and Q-learning are significantly affected by the complexity, while it is not the case for the IBL-Equal and IBL-Exponential model or humans in Experiment 1 (see ANOVA Table~\ref{tab:anova_redundancy}).
% The redundancy reduces sharply for the IBL-Equal and IBL-Exponential models compared to that of the TD models, which could only reach the zero redundancy level of humans at the very end. 
% Post-hoc Tukey's tests show that the mean redundancy of IBL-TD is significantly lower than Q-learning's redundancy ($p=.02$) in simple decisions but they do not differ in complex settings. 

Overall, the results display a main difference in humans' and models' behavior with respect to redundancy; the models make significantly more redundant actions than humans. The redundancy of human participants with restricted information in Experiment 2 and TD models was significantly influenced by the environment decision complexity, as opposed to that of human players in Experiment 1, IBL-Equal and IBL-Exponential whose redundancy across complexity levels remained unaffected.

% %------------
\subsubsection{Linear Movement Strategy}
Here we analyze the trajectories adopted by humans and the models to explain why the models made a lot more redundant visits early on while humans did not. 

Through our observation of human participants' trajectories, we noticed that some participants tended to go straight in a line of movement from their spawn location, rather than moving pivoting their trajectories. To look at whether such a strategy was indeed common in humans, and unlikely in the models, we compared a linear strategy, defined by making four consecutive steps in the same direction, i.e., either up, down, left, or right direction from the spawn location.

\begin{figure}[!htbp]
\centering
\includegraphics[width=1\linewidth]{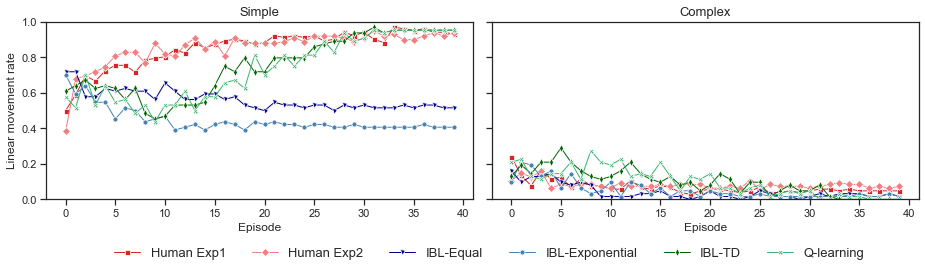}
\caption{The proportion that humans and agents made linear movement from spawn location in each condition.}
\label{fig:linear_movement}
\end{figure}

Fig.~\ref{fig:linear_movement} shows the proportion of linear movement by humans and models over episodes, in simple and complex decisions. It is clear from these results that decision complexity was a major factor in the adoption of a linear movement strategy for the models and humans. These main effects are confirmed by the statistical analyses reported in the Supplementary material, Table~\ref{tab:anova_linear_movement}. 

The effect of complexity in this case, however, can be explained by the design of complex decisions in the gridworlds, exemplified in Fig~\ref{fig:complex}. In complex decision gridworls, it is less likely that humans and models could move in a straight line from their spawn position because we designed these gridworlds so that a blocker would be located on the direct path towards the highest value target. That is, in complex decisions, agents were forced to explore more and break the trend of the linear movement strategy.
As a result, executing a linear strategy (i.e., four consecutive movements in a straight line) towards the highest value target was not possible in most of the cases when making complex decisions, and consequently the proportion of linear strategies is very low and decreases rapidly (see Supplementary Table~\ref{tab:linear_movement_stat} and Fig.~\ref{fig:linear_movement_2} for descriptive statistics).

In simple decision environments, humans follow the linear movement strategy from the initial episodes. By design, it is worth noting that the highest value targets are located in a linear path from the spawn location in simple decision environments. Although humans did not know where the highest value target was located, humans increasingly followed a linear strategy over episodes. This strategy did not differ in Experiment 1 compared to Experiment 2 (see Supplementary Table~\ref{tab:tukey_linear_movement}). There is no statistically significant difference in the proportion of the linear strategy between humans playing the task in Experiments 1 and 2 in simple or complex conditions. Thus, the linear strategy is not dependent on the amount of information provided to participants.

Moreover, in simple decision gridworlds, we did not find significant mean differences between human players in Experiment 2 and the TD models (see Supplementary Table~\ref{tab:tukey_linear_movement}). We further see that although the TD models' behavior does not reflect the linear strategy initially, the TD models are the only ones that can learn to match such behavior of humans in the last episodes (see Supplementary Fig.~\ref{fig:linear_movement_2}).

Finally, we see that, in contrast to humans and the TD models, the rate that the IBL-Equal and IBL-Exponential agents make linear movements in simple environments decreases during episodes and becomes stable in the last episodes. The explanation for this behavior of the IBL-Equal and IBL-Exponential agents is that they follow a non-linear path, and if the agents happen to find the highest value target, the path is reinforced. Consequently, they continue to follow such non-linear path, explaining why these models cannot account for humans' POptimal despite their success in doing so in terms of PMax. In contrast, the IBL-TD and Q-learning agents learn to estimate the value of each action at each state based on the difference between the expected and observed reward, without waiting for a final outcome. Over time, the estimate is propagated back and updated to be more accurate, resulting in the optimal performance (getting the highest value target with the shortest path) after a sufficient number of samplings (explorations). Thus, they were able to surpass the POptimal of humans, eventually. 

In summary, humans are significantly affected by the level of decision complexity; they are able to find a target in early episodes, by avoiding redundant decisions, particularly when given more information about the structure of the task; and they, at the outset, have a bias towards executing a linear strategy which is effective particularly in simple environments. By contrast, TD models most often end up reaching no target in the beginning since they make significantly more redundant visits, even though over time, they manage to learn the linear strategy, the optimal strategy for getting the highest value target in simple settings. IBL-Exponential and IBL-Equal are able to find a target quickly, but these strategies are not optimal.

% %------------
\subsection{Opportunities to Enhance Human Learning}
While the TD models might not learn as efficiently as humans do initially, they eventually surpass humans' performance in terms of PMax and POptimal as shown in figures Fig.~\ref{fig:default_parameter_learning_curve} and Fig.~\ref{fig:onestep_task}. The TD models become significantly more effective than humans, especially in complex environments and with limited information. How can we explain that despite all the inefficiencies above, TD agents turn out to be significantly more efficient than humans? We offer two explanations: excessive initial exploration and temptation avoidance, which reflect and are pertinent to human bias of believing that the low-hanging fruit is the ripest. 

\subsubsection{Initial Exploration and Decision Complexity} 
As explained in the previous section, the TD models perform very poorly initially as they often find no targets in their initial episodes, make highly redundant actions, and do not have any tendency to follow a linear strategy initially, as humans do. Here we analyze the coverage of the gridworld space, determined by the ratio of the number of uniquely visited locations to the total accessible locations. This metric specifies how much one explores the whole space.

\begin{figure}[!htbp]
\centering
\includegraphics[width=1\linewidth]{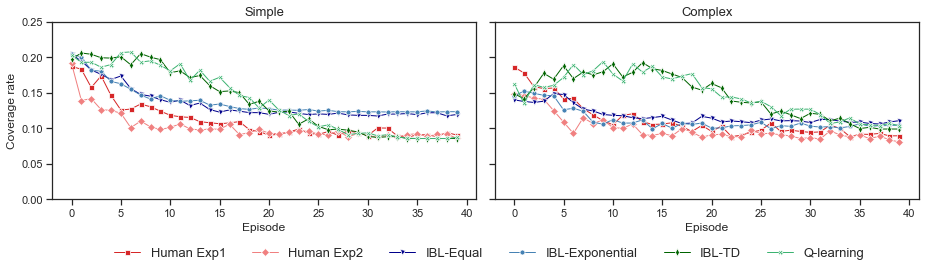}
\caption{The coverage rate of humans and the agents over time in each condition.}
\label{fig:coverage}
\end{figure}

Fig.~\ref{fig:coverage} provides the coverage of humans and models across episodes under the two decision complexity settings. Detailed statistical analyses can be found in the supplementary material, Table~\ref{tab:anova_coverage} and \ref{tab:tukey_coverage}.

Our first observation is that the initial coverage of IBL-TD and Q-learning models is much higher than that of humans and other models. Second, the decision complexity affects models' coverage, whereas humans' coverage is not affected by decision complexity. All the models explore less in complex environments than in simple ones, and more at the beginning than at the end.
ANOVA statistic results corroborate our observation that coverage by humans is remarkably similar across decision complexity levels (see ANOVA Table~\ref{tab:anova_coverage}). Human coverage is, however, significantly higher with more information (i.e., Experiment 1) than with limited information (i.e., Experiment 2) (see Supplementary Table~\ref{tab:tukey_coverage})

% %------------
\subsubsection{Temptation to Get Close Distractors.}
In our study, we recall that there is only one highest value target (\textit{preferred target}) while the other three act as distractors. One of the distractors is intentionally located close to the spawn location, and the difference in the distance to the closest distractor and the highest value target determines the complexity of the decision (see Fig.~\ref{fig:conditions}).

\begin{figure}[!htbp]
\centering
\includegraphics[width=1\linewidth]{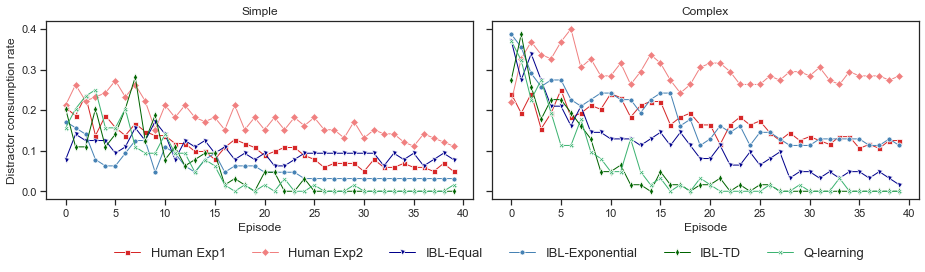}
\caption{The proportion humans and the agents that consumed the closest distractor over time in each condition.}
\label{fig:closest_distractor}
\end{figure}

Interestingly, we observe above that human exploration is not influenced by the decision complexity while the models' exploration is. Our conjecture is that humans cannot resist the temptation of earning the points of the closest distractor and would prefer to settle on the closest distractor instead of exploring more to find the highest value target.

%as the cost to make a movement affects people's behavior~\citep{hagura2017perceptual}.
To analyze this possibility, we looked at the proportion of times that humans and models reached the closest distractor during episodes. Fig.~\ref{fig:closest_distractor} depicts these proportions and a detailed statistical analysis is shown in the Supplementary Table~\ref{tab:anova_distractor} and ~\ref{tab:tukey_distractor}.

The results clearly show that humans consumed the closest distractor more often than models, in both simple and complex environments. Post-hoc comparisons using the Tukey HSD test indicate that the mean proportion that humans got the closest distractor under simple ($M=0.11$, $SD=0.31$) and complex settings ($M=0.17$, $SD=0.38$) in Experiment 1, and in Experiment 2 under simple ($M=0.18$, $SD=0.38$) and complex settings ($M=0.30$, $SD=0.45$), was significantly higher than the mean proportion IBL-TD or Q-learning agents consumed the distractor, regardless of the complexity levels (i.e., IBL-TD and Q-learning both have $M=0.06$ and $SD=0.24$ in the simple and complex settings) (see Supplementary Table~\ref{tab:tukey_distractor}).

The results also show a significant effect of the complexity on the consumption of the closest distractor for humans in Experiment 2 but not in Experiment 1. Also, humans in Experiment 2 consumed the closest distractor more often in Experiment 1. 
In terms of the models, no significant effect was found for the closest distractor consumption of IBL-Equal, IBL-TD and Q-learning agents (see ANOVA~\ref{tab:anova_distractor}).

In summary, our results show that models tend to explore more than humans initially and they are able to avoid the temptation of close distractors better than humans do. These two aspects of the models represent potential opportunities to influence humans to become wider explorers and to avoid their temptation to handle decision complexity more effectively.

\section{Discussion and Conclusions}
Current research in Artificial Intelligence (AI) addresses the temporal credit assignment in sequential decision making tasks mostly by using the temporal difference (TD) learning method in reinforcement learning (RL) models (Q-learning). This method has been very successful in dealing with problems in which feedback is delayed. The TD algorithms enable AI agents to reach levels of human performance in many complex decision making tasks \citep{wong2021multiagent}. Although RL models are able to solve computational problems efficiently and reach high-performance levels, it is unclear how the TD approach is aligned with the actual human behavior \citep{BOTVINICK19,lake2017building}. In this paper, we investigated three different mechanisms of credit assignment (Equal, Exponential and TD) using the cognitive models, constructed based on IBLT \citep{GONZALEZ03}, a cognitive theory of decisions from experience.  Simulation results from IBL models and a Q-learning model were compared to human data obtained in a goal-seeking navigation task equipped with various levels of decision complexity, which is particularly well suited for studying the temporal credit assignment problem.

Overall, we elucidate the differences in how these models learn with delayed feedback compared to humans. Furthermore, we also find ways in which the models are able to outperform humans. Understanding the reasons behind these differences will help to inform the development of human-like AI agents, and agents that can also successfully support and help improve human decisions and learning.

We find that Equal and Exponential credit assignment mechanisms can capture initial human learning and generally, the shape of human learning curves in terms of PMax (i.e. the ability to find the highest value target). However, these credit assignment mechanisms show inefficient behavior compared to humans in terms of POptimal (i.e. the ability to reach the highest value target via the shortest path), especially in simple decision environments. These results suggest that humans might be initially unable to predict the value of the immediate states, without having received feedback; rather they update the value of their initial actions once they have received feedback at the end of an episode. Furthermore, it is also possible that even after receiving outcome feedback, humans are initially unable to distinguish how the various decisions made throughout an episode actually contributed to the final outcome. Thus, crediting all the actions equally is a simple procedure that humans may be adopting in the initial learning.  This interpretation is further supported by the fact that the Equal and Exponential mechanisms show sensitivity to decision complexity, similar to humans, but the TD models are unaffected by decision complexity. 

The TD mechanisms learn much slower than humans early in the tasks but then achieve the same level of human optimal performance and become more efficient than humans later in the task. The detailed analyses of the process metrics reveal that the TD mechanisms lead to more initial exploration of the environment compared to humans. Such larger exploration leads the TD models to a higher coverage of the environment, and consequently to a better ``understanding'' of where obstacles and targets are located. As the TD models learn more about the environment, they also become more efficient in reaching the highest value target in the shortest path, and ultimately they do this better than humans. In attempting to fully explore the environment, the TD models also made far more redundant visits initially, compared to humans; which resulted in a greater proportion of episodes in which no targets were found by the model, compared to humans during the initial exploration. The major implication of this result is that we can use help humans to engage in more initial exploration; which would help humans to learn faster and to become ultimately more optimal and accurate. 

Our results suggest two reasons that prevent humans from exploring more initially and from being as optimal as the TD models are with extended practice.  First, humans showed a strong tendency to make a linear movement while navigating the environment, preventing them from further exploration. This is indicative of strategies that humans might have learned from experience when approaching unknown problems while trying to be efficient; that is, humans do not learn each task from scratch as models do ~\citep{pouncy2021model,firestone2020performance}. Second, humans have a strong tendency to be tempted to earn the points offered by the closest distractor instead of exploring to find the highest value target. To some extent, such characteristic is in alignment with a very well-known ``satisficing'' human behavior in a decision making process; where people often select an option that is good enough, and not necessarily optimal~\citep{simon1956rational}. We plan to investigate these costs (i.e., distance to the target) and benefits (i.e., values of the target) tradeoffs in future research.

Ultimately, biases in human behavior present an opportunity for developing AI agents that can help humans improve their decisions.  
The implication of our findings highlights the importance of exploration to make the right decision about the targets. Lacking adequate exploration, it is more likely that humans make decisions based on imperfect information about the environment, leading to sub-optimal decisions (i.e. the higher rate of consuming the closest distractor). Indeed, this is in agreement with research on the exploration-exploitation tradeoff in humans, where with more experience and exploration humans learn to improve their decision optimality \citep{mehlhorn2015unpacking}.
Presumably, a TD agent that has learned an effective strategy for a task after sufficient training can serve as a ``teacher'' or human tutor, who advises humans on which actions to take. It is similarly important to define when is the right time to intervene since over-advising could disrupt and hamper humans' learning~\citep{torrey2013teaching,da2020uncertainty}.
The IBL-TD model may be particularly amenable to determine when to provide advice to humans, given that the IBL model may be able to characterize the ``state of mind'' by populating humans' experience in the model's memory~\citep{nguyen2021theory}, while also including the credit assignment power of the TD model. RL research predominantly focused on AI-AI advisor-advisee or on improving AI through human intervention~\citep{torrey2013teaching,griffith2013policy} rather than improving human decisions through AI interventions. We consider this direction worthy of investigation.

In summary, the findings of this research reveal how little is known regarding human credit assignment to decisions when feedback is delayed.  Initially, in learning a new task, human decisions are more in agreement with Equal credit assignment, suggesting that humans might wait until obtaining an outcome after a sequence of actions, and then assign equal credit to all the actions in the sequence. The widely-used TD approach for credit assignment does not represent the initial learning of humans in a task. As proposed by research in model-based RL, this challenge can be alleviated by pre-training a model or by integrating some rule-based planning approaches that can handle the initial knowledge carried by humans from experience ~\citep{daw2011model,otto2013curse,pouncy2021model}. These approaches, however, do not address the major question of how humans handle the credit assignment problem early on when confronting a new task.

Finally, future research should focus on an explicit investigation of the TD credit assignment approach and how to compare and help humans adapt in the presence of feedback delays, rather than inferring the credit assignment mechanism from human behavior and the comparison to model behavior. Future research should also investigate the use of cognitive models as humans' teammates in the context of human-machine teaming since we anticipate that humans will be able to infer and predict the actions of cognitive IBL agents better than those of other, non-cognitive models. We leave this demonstration for future research.

%Acknowledgments
\section{Acknowledgements}
This research was supported by the Defense Advanced Research Projects Agency and was accomplished under Grant Number W911NF-20-1-0006; and by the AFRL Award FA8650-20-F-6212 and sub-award number 1990692.
% \input{7_method.tex}
% \bibliography{reference}

%\input{10_supplementary}
\clearpage
% \begin{appendices}
%%%%%%%%%% Merge with supplemental materials %%%%%%%%%%
\pagebreak
\begin{center}
\textbf{\huge Supplementary Materials}
\end{center}
%%%%%%%%%% Merge with supplemental materials %%%%%%%%%%
%%%%%%%%%% Prefix a "S" to all equations, figures, tables and reset the counter %%%%%%%%%%
\setcounter{section}{0}
\setcounter{equation}{0}
\setcounter{figure}{0}
\setcounter{table}{0}
\setcounter{page}{1}
\makeatletter
\renewcommand{\theequation}{S\arabic{equation}}
\renewcommand{\thetable}{S\arabic{table}}
\renewcommand{\thefigure}{S\arabic{figure}}
\renewcommand{\thesection}{S-\Roman{section}}
% \renewcommand{\bibnumfmt}[1]{[S#1]}
% \renewcommand{\citenumfont}[1]{S#1}
%%%%%%%%%% Prefix a "S" to all equations, figures, tables and reset the counter %%%%%%%%%%

% %---------------------------------
% %-----Experiment screenshots------
% %---------------------------------
\section{Instructions} \label{sec:instructions}
\begin{figure}[!htbp]
\centering
\includegraphics[width=0.61\linewidth]{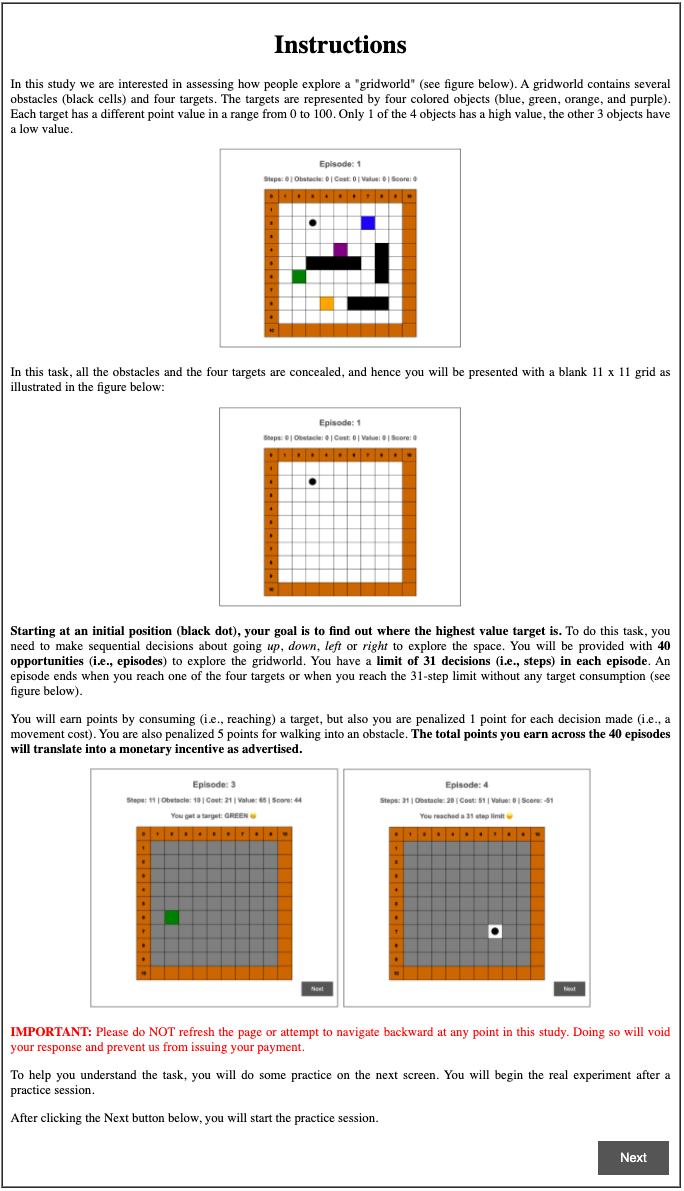}
\caption{Instructions presented in the Experiment 1.}
\label{fig:instruction_1}
\end{figure}

\begin{figure}[!htbp]
\centering
\includegraphics[width=0.8\linewidth]{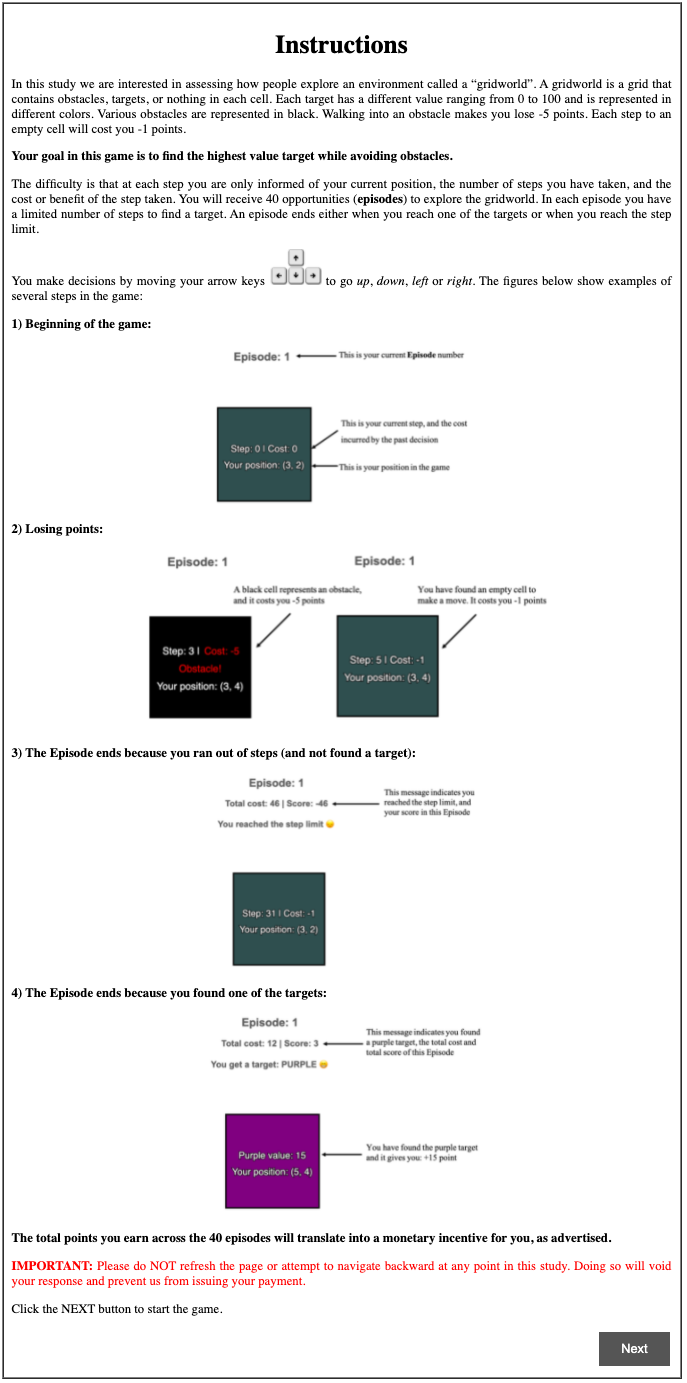}
\caption{Instructions presented in the Experiment 2.}
\label{fig:instruction_2}
\end{figure}

Fig.~\ref{fig:instruction_1} and \ref{fig:instruction_2} are screenshots for the instruction of the game presented to participants in Experiment 1 and 2, respectively. 

% %---------------------------------
% %--------Human Fitting-----------
% %---------------------------------

%\input{5b_human_fit}
\section{Fitting the models' parameters to behavioral data} \label{sec:human_fit}
As a quick exercise, to try to assess whether fitting the models' parameters to human data would improve the way the main effects observed from pure predictions of the models, we fit each of the models' parameters to humans' average PMax separately in the two experiments and for each complexity condition. We conducted a search over all parameters for each model with the goal of minimizing the RMSE between the model's PMax and the human PMax at each episode. The search space for all parameters for all models was $(0,1)$. The resulting parameters are summarized in Table \ref{tab:human_fit_params}.

\begin{table}[!htpb]
\resizebox{.9\textwidth}{!}{% 
\begin{tabular}{@{}llllllllllll@{}}
\toprule
 &  & \multicolumn{5}{c}{Simple} & \multicolumn{5}{c}{Complex} \\ \cmidrule(l){3-12} 
 & Model & $\sigma$ & $d$ & $\gamma$ & $\alpha$ & $\epsilon$ & $\sigma$ & $d$ & $\gamma$ & $\alpha$ & $\epsilon$ \\ \midrule
\multirow{4}{*}{Experiment 1} & IBL-Equal & 0.814 & 0.143 & --- & --- & --- & 0.014 & 0.400 & --- & --- & --- \\
 & IBL-Exponential & 0.162 & 0.800 & 0.998 & --- & --- & 0.269 & 0.999 & 0.980 & --- & --- \\
 & IBL-TD & 0.065 & 0.484 & 0.966 & 0.898 & --- & 0.034 & 0.969 & 0.730 & 0.878 & --- \\
 & Q-learning & --- & --- & 0.997 & 0.839 & 0.002 & --- & --- & 0.954 & 0.714 & 0.020 \\ \midrule
\multirow{4}{*}{Experiment 2} & IBL-Equal & 0.716 & 0.057 & --- & --- & --- & 0.029 & 0.682 & --- & --- & --- \\
 & IBL-Exponential & 0.012 & 0.121 & 0.943 & --- & --- & 0.047 & 0.243 & 0.865 & --- & --- \\
 & IBL-TD & 0.014 & 0.862 & 0.576 & 0.971 & --- & 0.909 & 0.359 & 0.776 & 0.494 & --- \\
 & Q-learning & --- & --- & 0.548 & 0.978 & 0.03 & --- & --- & 0.576 & 0.722 & 0.310 \\ \bottomrule
\end{tabular}
}
\caption{Parameters resulting from the calibration of the models to human data in Experiment 1 and in Experiment 2.}
\label{tab:human_fit_params}
\end{table}

Table~\ref{tab:exp_fit} shows the resulting average PMax performance of the models and the difference to human average Pmax for Experiment 1 and Experiment 2. As expected, calibrating the models' parameters to human PMax improves the similarity of the models to human PMax in both experiments. 

\begin{table}[!htpb]
\centering
\resizebox{.6\textwidth}{!}{% 
% \input{tables/Tab_Exp1Results_Fit}
% Combined
\begin{tabular}{llllll}
\hline
 & \multirow{2}{*}{Decision Complexity} & \multicolumn{2}{c}{Simple} & \multicolumn{2}{c}{Complex} \\ \cline{3-6} 
 &  & \begin{tabular}[c]{@{}l@{}}PMax\\ Avg.\end{tabular} & \begin{tabular}[c]{@{}l@{}}PMax\\ Err.\end{tabular} & \begin{tabular}[c]{@{}l@{}}PMax\\ Avg.\end{tabular} & \begin{tabular}[c]{@{}l@{}}PMax\\ Err.\end{tabular} \\ \hline
 \multirow{5}{*}{Experiment 1} & Human  & 0.71 & -- & 0.48 & --- \\ \cdashlinelr{1-6}
& IBL-Equal       & 0.77 & 0.06   & 0.60 &  0.12  \\
& IBL-Exponential & 0.74 & 0.03   & 0.55 &  0.07  \\
& IBL-TD          & 0.66 & -0.05   & 0.44 &  -0.04  \\
& Q-learning      & 0.67 & -0.04   & 0.43 &  -0.05  \\ \hline
\multirow{5}{*}{Experiment 2}  & Human           &   0.63 & ---    & 0.24 & ---    \\ \cdashlinelr{1-6}
& IBL-Equal       &   0.62 &  -0.01  & 0.47 &  0.23  \\
& IBL-Exponential &   0.61 &  -0.02  & 0.31 &  0.07  \\
& IBL-TD          &   0.63 &  0.00  & 0.27 &  0.03  \\
& Q-learning      &   0.60 &  -0.03  & 0.25 &  0.01  \\ \bottomrule
\end{tabular}
}
\caption{Average performance and error for models fit to human data from Experiment 1 and Experiment 2.}
\label{tab:exp_fit}
\end{table}

\begin{figure}[!htpb]
    \centering
    \begin{subfigure}[b]{\textwidth}
        \centering
         \includegraphics[width=.9\textwidth]{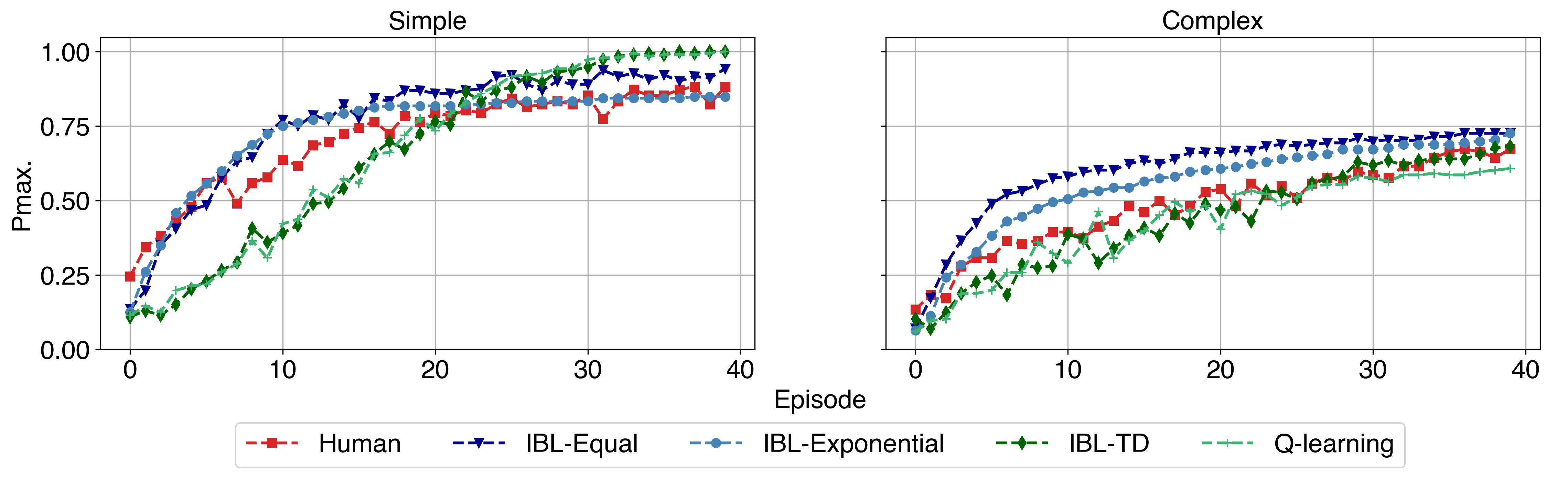}
         \caption{Experiment 1}
         \label{fig:exp1_fit_curve}
    \end{subfigure}
    \begin{subfigure}[b]{\textwidth}
        \centering
         \includegraphics[width=.9\textwidth]{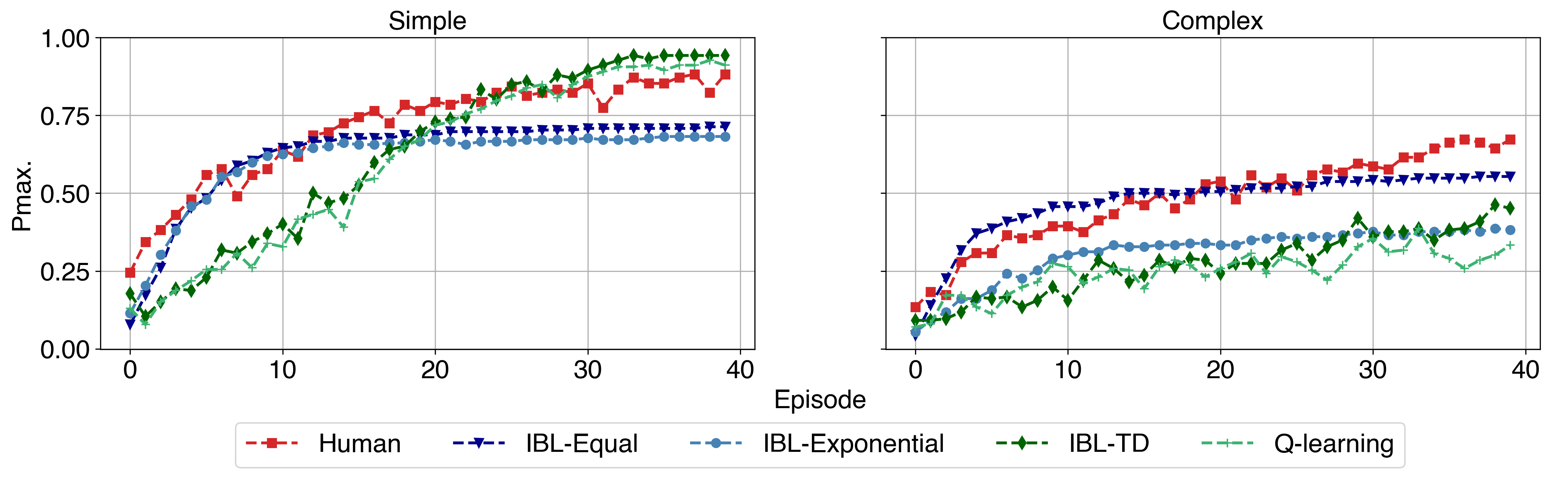}
         \caption{Experiment 2}
         \label{fig:exp2_fit_curve}
    \end{subfigure}
    % % \resizebox{.99\textwidth}{!}{% 
    % % \includegraphics[width=\textwidth]{fig/LearningCurvePlots/EXP1_Fit_PMax.png}
    % % }
    \caption{PMax results with parameters selected to minimize the average difference between human (a) Experiment 1 and (b) Experiment 2 and model PMax.}
    \label{fig:exp_fit}
\end{figure}

Figure~\ref{fig:exp_fit} shows the resulting PMax per episode, for the models calibrated to human data in Experiment 1 and Experiment 2. We observe some improvement in capturing the PMax per episode in some models in both experiments. However, the TD models, even after fitting their parameters, are unable to capture the initial PMax of human participants, especially in simple environments.

\section{Statistical analysis~\protect\footnote{All analyses were performed using R version 3.5.3.}} \label{sec:Statistics}

A mixed ANOVA was conducted to compare the main effects of decision complexity levels (simple and complex) and  episodes (40 episodes) as well as the interaction effect between them, on the behavioral characteristics of humans and the models with regard to each of the following process metrics.  
We further conducted an ANOVA test to compare humans' and models' behavior in each decision complexity condition, which was performed separately for each condition. 
Finally we provide descriptive statistics for the first 10 and the last 10 playing episodes in terms of the corresponding metrics.

% %---------------------------------
% %-------Target consumption--------
% %---------------------------------
\subsection{Target consumption}
Table~\ref{tab:anova_target_consumption} reports the ANOVA for the effects on humans' and the models' each target and no target consumption rate. The ANOVA was performed separately for each of them. 

\begin{table}[H]
% \centering
\resizebox{.93\textwidth}{!}{% 
\begin{tabular}{@{}llllrlrlrlrlrl@{}}
\toprule
\multirow{2}{*}{} & \multirow{2}{*}{Effect} & \multirow{2}{*}{DFn} & \multirow{2}{*}{DFd} & \multicolumn{2}{c}{1st highest} & \multicolumn{2}{c}{2nd highest} & \multicolumn{2}{c}{3rd highest} & \multicolumn{2}{c}{4th highest} & \multicolumn{2}{c}{No target} \\ \cmidrule(l){5-14} 
 &  &  &  & \multicolumn{1}{c}{F} & \multicolumn{1}{c}{p} & \multicolumn{1}{c}{F} & \multicolumn{1}{c}{p} & \multicolumn{1}{c}{F} & \multicolumn{1}{c}{p} & \multicolumn{1}{c}{F} & \multicolumn{1}{c}{p} & \multicolumn{1}{c}{F} & \multicolumn{1}{c}{p} \\ \midrule
\multirow{3}{*}{{\begin{tabular}[c]{@{}l@{}}Human Experiment 1 \end{tabular}}} & complexity & 1 & 204 & 25.04 & \textbf{0.000}  &  3.84 & 0.052 & 10.67 &\textbf{0.001} & 4.21 & \textbf{0.042} & 4.81 & \textbf{0.029} \\
 & episode & 39 & 7956 & 47.84 & \textbf{0.000} & 3.96 & \textbf{0.000} & 6.53 & \textbf{0.000} & 9.24 & \textbf{0.000} & 8.82  & \textbf{0.000} \\
 & complexity:episode & 39 & 7956 & 1.19 & 0.192  & 1.27 & 0.122 & 1.02 & 0.435 & 1.19 & 0.189 & 0.95 & 0.568 \\ \hline
 
 \multirow{3}{*}{{\begin{tabular}[c]{@{}l@{}}Human Experiment 2 \end{tabular}}} & complexity & 1 & 192 & 59.98 &  \textbf{0.000}& 6.29 & \textbf{0.013} & 17.92 & \textbf{0.000} & 7.12 & \textbf{0.008} & 4.71 & \textbf{0.031} \\
 & episode & 39 & 7488 & 21.50 & \textbf{0.000} & 0.86 & 0.720 & 2.30 & \textbf{0.000} & 4.81  & \textbf{0.000} & 11.63 & \textbf{0.000} \\
 & complexity:episode & 39 & 7488 & 3.25 & \textbf{0.000} & 3.47 & \textbf{0.000} & 1.01 & 0.459 & 0.86 & 0.710 & 1.19 & 0.191 \\ \hline
 
 \multirow{3}{*}{IBL-Equal} & complexity & 1 & 124 & 3.70 & 0.056 & 1.42 & 0.236 & 9.91 & \textbf{0.002} & 7.20 & \textbf{0.008} & 9.73  & \textbf{0.002}\\
 & episode & 39 & 4836 & 93.76 & \textbf{0.000} & 0.85 & 0.736  & 6.27 & \textbf{0.000} & 7.78 & \textbf{0.000} & 41.19 & \textbf{0.000} \\
 & complexity:episode & 39 & 4836 & 1.57 & \textbf{0.014} & 0.71 & 0.916  & 3.53  & \textbf{0.000} & 2.08 & \textbf{0.000}  & 5.10 & \textbf{0.000} \\ \hline
 
 \multirow{3}{*}{IBL-Exponential} & complexity & 1 & 124 & 11.45 & \textbf{0.001} & 4.66 & \textbf{0.033} & 9.7 & \textbf{0.002} & 6.47 & \textbf{0.012} & 14.00 & \textbf{0.000} \\
 & episode & 39 & 4836 & 97.08 & \textbf{0.000} & 2.45 & \textbf{0.000} & 3.88 & \textbf{0.000} & 7.59 & \textbf{0.000} & 43.02 & \textbf{0.000} \\
 & complexity:episode & 39 & 4836 & 0.94 & 0.577 & 0.38 & 0.999  & 0.95 & 0.564 & 2.60 & \textbf{0.000} & 6.59 & \textbf{0.000} \\ \hline
 
 \multirow{3}{*}{IBL-TD} & complexity & 1 & 124 & 7.05 & \textbf{0.009} & 3.49 & 0.064  & 25.54 & \textbf{0.000} & 5.72 & \textbf{0.018} & 0.10 & 0.755 \\
 & episode & 39 & 4836 & 101.93 & \textbf{0.000} & 7.38 & \textbf{0.000} & 8.19 & \textbf{0.000} & 9.82 & \textbf{0.000} & 25.10 & \textbf{0.000} \\
 & complexity:episode & 39 & 4836 & 1.52 & \textbf{0.021} & 1.35 & 0.0721 & 2.16 & \textbf{0.000} & 2.66 & \textbf{0.000} & 4.94 & \textbf{0.000} \\ \hline
 
 \multirow{3}{*}{Q-Learning} & complexity & 1 & 124 & 5.86 & \textbf{0.017} & 4.28 & \textbf{0.041}  & 19.64 & \textbf{0.000} & 6.80 & \textbf{0.010} & 0.005 & 0.944  \\
 & episode & 39 & 4836 & 89.13 & \textbf{0.000} & 6.67  & \textbf{0.000} & 6.42 & \textbf{0.000}  & 8.06 & \textbf{0.000} & 24.21 & \textbf{0.000} \\
 & complexity:episode & 39 & 4836 & 1.15 & 0.246 & 0.89 & 0.659  & 1.53 & \textbf{0.019}  & 1.51  & \textbf{0.022} & 3.20 & \textbf{0.000} \\ \bottomrule
\end{tabular}
}
\caption{ANOVA for the effect of decision complexity on the humans and models' target consumption rate.}
\label{tab:anova_target_consumption}
\end{table}

Table~\ref{tab:tukey_consumption} presents the comparison between the mean proportion of times that the highest value target and no target is consumed by humans and the models.
\begin{table}[H]
\resizebox{.99\textwidth}{!}{% 
\begin{tabular}{@{}lllllllllll@{}}
\toprule
 &  & \multicolumn{4}{c}{1st highest value target} &  & \multicolumn{4}{c}{No target} \\ \cmidrule(l){3-11} 
\multicolumn{2}{l}{} &  \multicolumn{2}{c}{Simple} & \multicolumn{2}{c}{Complex} &  & \multicolumn{2}{c}{Simple} & \multicolumn{2}{c}{Complex} \\ \cmidrule(l){3-11} 
Group 1 & Group 2 & \multicolumn{1}{c}{diff} & \multicolumn{1}{c}{p.adj} & \multicolumn{1}{c}{diff} & \multicolumn{1}{c}{p.adj} &  & \multicolumn{1}{c}{diff} & \multicolumn{1}{c}{p.adj} & \multicolumn{1}{c}{diff} & \multicolumn{1}{c}{p.adj} \\ \midrule
Human Experiment 2 & Human Experiment 1 &  -0.084  & 0.000  & -0.245  & 0.000 & &  0.028  & 0.000  &  0.036 & 0.000  \\ 
IBL-Exponential & Human Experiment 1 & -0.664 & 0.000  & -0.358 & 0.000 & &  0.793  & 0.000  &   0.597  & 0.000  \\
IBL-Equal &  Human Experiment 1 & -0.633 &  0.000 &  -0.353  & 0.000   & &  0.754  & 0.000  & 0.632  & 0.000  \\ 
IBL-TD & Human Experiment 1 & -0.666  & 0.000  &  -0.421 &  0.000  & &  0.655  & 0.000  &   0.571  & 0.000  \\
Q-learning & Human Experiment 1 &  -0.668  & 0.000  &   -0.423  & 0.000   & &  0.650  & 0.000  &   0.572  & 0.000 \\
IBL-Exponential & Human Experiment 2 &  -0.581  & 0.000  &  -0.113 &  0.000   & &   0.765 & 0.000   &  0.561  & 0.000  \\ 
IBL-Equal & Human Experiment 2 &  -0.549 & 0.000  &  -0.108 &  0.000   & &  0.726 & 0.000   &   0.595  & 0.000 \\ 
IBL-TD & Human Experiment 2 &  -0.582 & 0.000 & -0.177  & 0.000   & &  0.627  & 0.000   &   0.535 &  0.000  \\ 
Q-learning & Human Experiment 2 & -0.585  & 0.000 &  -0.178  & 0.000   & &  0.623  & 0.000  &  0.536  & 0.000 \\ 
IBL-Exponential & IBL-Equal &  -0.032  & 0.012  &  -0.005 & 0.997  & &  0.039  & 0.000   &  -0.034  & 0.006   \\ 
IBL-TD & IBL-Exponential &  -0.002  & 1.000  &  -0.064  & 0.000  & &   -0.138  & 0.000   &   -0.026  & 0.080 \\ 
Q-learning & IBL-Exponential &  -0.004 & 0.999  & -0.065 & 0.000  & &   -0.143 & 0.000  &    -0.025 & 0.099   \\ 
IBL-TD & IBL-Equal &  -0.033  & 0.007  &  -0.069 &  0.000 & &  -0.098 & 0.000  &   -0.060 &  0.000 \\ 
Q-learning & IBL-Equal &  -0.036  & 0.003  &-0.070  & 0.000  & & -0.103  & 0.000  &  -0.060  & 0.000  \\ 
Q-learning & IBL-TD &  -0.002  & 1.000  &  -0.002  & 1.000  & & -0.005  & 0.993 &   0.001 & 1.000   \\ \bottomrule
\end{tabular}
}
\caption{ANOVA with post-hoc Tukey HSD for comparing the mean proportion of times that the \textbf{highest value target} and \textbf{no target} is obtained by humans and the models.}
\label{tab:tukey_consumption}
\end{table}

% %---------------------------------
% %-------Redundancy--------
% %---------------------------------
\subsection{Redundancy and Immediate redundancy}
Table~\ref{tab:anova_redundancy} summarizes a mixed ANOVA for the effects of decision complexity on the redundancy and the immediate redundancy.

Table~\ref{tab:tukey_redundancy} reports the comparison among humans and the models in terms of the average proportion of making redundant and immediately redundant visits. 

\begin{figure}[!htbp]
\centering
\includegraphics[width=0.93\linewidth]{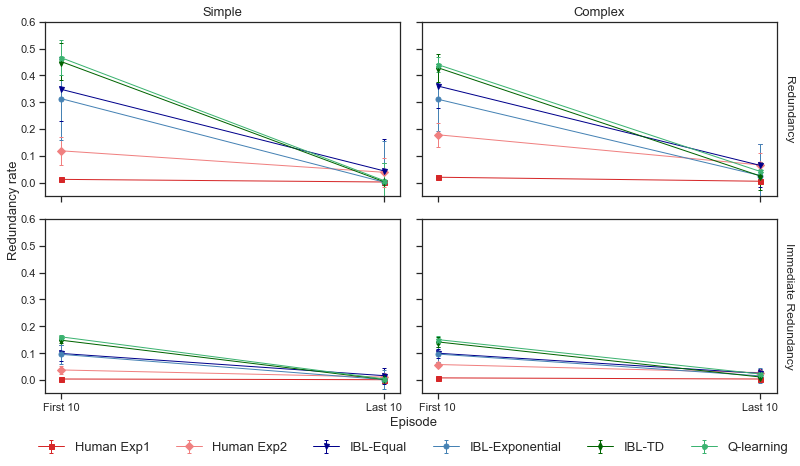}
\caption{The average \textbf{redundancy} and \textbf{immediate redundancy} rate in the first 10 and last 10 episodes.}
\label{fig:redundancy_2}
\end{figure}

Table~\ref{tab:redundancy_stat} and Fig.~\ref{fig:redundancy_2} display the average redundancy and immediate redundancy rate in the first and last 10 episodes.

\begin{table}[H]
\resizebox{.9\textwidth}{!}{% 
\begin{tabular}{@{}llllrrrr@{}}
\toprule
\multirow{2}{*}{} &  &  &  & \multicolumn{2}{r}{Redundancy} & \multicolumn{2}{c}{\begin{tabular}[c]{@{}c@{}}Immediate \\ Redundancy\end{tabular}} \\ \cmidrule(l){5-8} 
 & Effect & \multicolumn{1}{c}{DFn} & \multicolumn{1}{c}{DFd} & \multicolumn{1}{c}{F} & \multicolumn{1}{c}{p} & \multicolumn{1}{c}{F} & \multicolumn{1}{c}{p} \\ \midrule
\multirow{3}{*}{Human Experiment 1} & complexity & 1 & 204 & 3.28 & 0.072  &  3.06 & 0.082  \\
 & episode & 39 & 7956 & 3.77 & \textbf{0.000}  &  1.56 & \textbf{0.014}  \\
 & complexity:episode & 39 & 7956 & 1.25 & 0.139  &  1.06 & 0.376  \\ \midrule
\multirow{3}{*}{Human Experiment 2} & complexity & 1 & 192 & 7.43 & \textbf{0.007}   &  5.24 & \textbf{0.023}  \\
 & episode &  39 & 7488 & 23.54 & \textbf{0.000}  &  6.56 & \textbf{0.000}  \\
 & complexity:episode &  39 & 7488 & 1.06 & 0.369  &  0.76 & 0.856  \\ \midrule
\multirow{3}{*}{IBL-Equal} & complexity &  1 & 124 & 3.27 & 0.073   &  1.75 & 0.188 \\
 & episode &  39 & 4836 & 91.31 & \textbf{0.000}  &  47.06 & \textbf{0.000}  \\
 & complexity:episode & 39.00 & 4836 & 1.63 & \textbf{0.008}  &  1.38 & 0.058  \\ \midrule
\multirow{3}{*}{IBL-Exponential} & complexity & 1 & 124 & 2.66 & 0.105  &  2.93 & 0.089   \\
 & episode &  39 & 4836 & 105.38 & \textbf{0.000}  &  58.32 & \textbf{0.000}  \\
 & complexity:episode &  39 & 4836 & 1.37 & 0.064  &  1.25 & 0.141  \\ \midrule
\multirow{3}{*}{IBL-TD} & complexity & 1 & 124 & 6.32 & \textbf{0.013}  &  8.33 & \textbf{0.005}   \\
 & episode &  39 & 4836 & 133.61 & \textbf{0.000}  &  90.14 & \textbf{0.000}  \\
 & complexity:episode &  39 & 4836 & 2.66 & \textbf{0.000} &  2.44 & \textbf{0.000}  \\ \midrule
\multirow{3}{*}{Q-learning} & complexity &  1 & 124 & 3.34 & \textbf{0.049}  &  6.96 & \textbf{0.009}   \\
 & episode &  39 & 4836 & 132.28 & \textbf{0.000} &  87.47 & \textbf{0.000}  \\
 & complexity:episode &  39 & 4836 & 2.31 & \textbf{0.000} &  1.95 & \textbf{0.000}  \\ \bottomrule
\end{tabular}
}
\caption{ANOVA for the effect of decision complexity on the  \textbf{redundancy} and \textbf{immediate redundancy} of humans and the models.}
\label{tab:anova_redundancy}
\end{table}

\begin{table}[H]
\resizebox{.99\textwidth}{!}{% 
\begin{tabular}{@{}lllllllllll@{}}
\toprule
 &  & \multicolumn{4}{c}{Redundancy} &  & \multicolumn{4}{c}{Immediate Redundancy} \\ \cmidrule(l){3-11} 
\multicolumn{2}{l}{} &  \multicolumn{2}{c}{Simple} & \multicolumn{2}{c}{Complex} &  & \multicolumn{2}{c}{Simple} & \multicolumn{2}{c}{Complex} \\ \cmidrule(l){3-11} 
Group 1 & Group 2 & \multicolumn{1}{c}{diff} & \multicolumn{1}{c}{p.adj} & \multicolumn{1}{c}{diff} & \multicolumn{1}{c}{p.adj} &  & \multicolumn{1}{c}{diff} & \multicolumn{1}{c}{p.adj} & \multicolumn{1}{c}{diff} & \multicolumn{1}{c}{p.adj} \\ \midrule
Human Experiment 2 & Human Experiment 1 &  0.058 & 0.000 &  0.091  & 0.000 & & 0.018 & 0.000 & 0.031 & 0.000 \\ 
IBL-Exponential & Human Experiment 1 & 0.093 & 0.000 &  0.108  & 0.000 & & 0.031 & 0.000 & 0.036 & 0.000\\
IBL-Equal &  Human Experiment 1 & 0.134  & 0.000 &  0.161 &  0.000 & & 0.041 & 0.000 & 0.045 & 0.000 \\ 
IBL-TD & Human Experiment 1 & 0.186  & 0.000 & 0.208 & 0.000 & & 0.070 & 0.000 & 0.079  & 0.000 \\
Q-learning & Human Experiment 1 & 0.199 & 0.000 &  0.213 &  0.000 & & 0.078 & 0.000 & 0.086  & 0.000 \\
IBL-Exponential & Human Experiment 2 & 0.035 & 0.000 & 0.017 &  0.003 & & 0.013  & 0.000 & 0.004  & 0.241 \\ 
IBL-Equal & Human Experiment 2 & 0.076 & 0.000 & 0.070 & 0.000  & & 0.023 & 0.000 & 0.014  & 0.000\\ 
IBL-TD & Human Experiment 2 & 0.127 & 0.000 & 0.116 &  0.000 & & 0.052  & 0.000 & 0.048  & 0.000 \\ 
Q-learning & Human Experiment 2 & 0.141 & 0.000 & 0.122 &  0.000 & & 0.059 & 0.000 & 0.055 & 0.000\\ 
IBL-Equal & IBL-Exponential & 0.041  & 0.000 & 0.053 & 0.000 & & 0.010  & 0.000  & 0.010 & 0.000 \\ 
IBL-TD & IBL-Exponential & 0.092 & 0.000 & 0.100 & 0.000 & & 0.039  & 0.000  & 0.044  & 0.000 \\ 
Q-learning & IBL-Exponential & 0.106 & 0.000 & 0.105 & 0.000 & & 0.047  & 0.000 & 0.050  & 0.000 \\ 
IBL-TD & IBL-Equal & 0.051 & 0.000 & 0.046 & 0.000 & & 0.029 & 0.000 & 0.034  & 0.000\\ 
Q-learning & IBL-Equal & 0.065 & 0.000 & 0.052  & 0.000 & & 0.037  & 0.000 & 0.041 & 0.000 \\ 
Q-learning & IBL-TD & 0.013 & 0.015 & 0.006 & 0.850 & & 0.008  & 0.000 & 0.007  & 0.034  \\ \bottomrule
\end{tabular}
}
\caption{ANOVA with post-hoc Tukey HSD for comparing the mean \textbf{redundancy} and \textbf{immediate redundancy} rate among humans and the models.}
\label{tab:tukey_redundancy}
\end{table}

% % %descriptive statistics
\begin{table}[H]
\resizebox{.9\textwidth}{!}{% 
\begin{tabular}{@{}lllllll@{}}
\toprule
\multicolumn{1}{c}{} & \multicolumn{2}{l}{\multirow{2}{*}{Humans and Models}} & \multicolumn{2}{c}{Simple} & \multicolumn{2}{c}{Complex} \\ \cmidrule(l){4-7} 
\multicolumn{1}{c}{} & \multicolumn{2}{l}{} & First 10 episodes & Last 10 episodes & First 10 episodes & Last 10 episodes \\ \midrule
\multirow{6}{*}{Redundancy}&\multicolumn{2}{l}{Human Experiment 1} & 0.012 $\pm$ 0.00 & 0.002 $\pm$ 0.00 & 0.020 $\pm$ 0.00 & 0.005 $\pm$ 0.00 \\
&\multicolumn{2}{l}{Human Experiment 2} & 0.119 $\pm$ 0.05   & 0.039 $\pm$ 0.05  & 0.179 $\pm$ 0.05 & 0.066 $\pm$ 0.05 \\
&\multicolumn{2}{l}{IBL-Equal} & 0.348 $\pm$ 0.12 & 0.044 $\pm$ 0.12 & 0.360 $\pm$ 0.08 & 0.064 $\pm$ 0.08\\
&\multicolumn{2}{l}{IBL-Exponential} & 0.313 $\pm$ 0.15 & 0.002 $\pm$ 0.15 & 0.311 $\pm$ 0.12 & 0.027 $\pm$ 0.12\\
&\multicolumn{2}{l}{IBL-TD} & 0.451 $\pm$ 0.07  & 0.004 $\pm$ 0.07 & 0.429 $\pm$ 0.05  & 0.025 $\pm$ 0.05\\ 
&\multicolumn{2}{l}{Q-Learning} & 0.466 $\pm$ 0.06  & 0.007 $\pm$ 0.06 & 0.440 $\pm$ 0.03  & 0.042 $\pm$ 0.03\\\midrule
\multirow{6}{*}{\begin{tabular}[c]{@{}l@{}}Immediate\\ Redundancy\end{tabular}} &\multicolumn{2}{l}{Human Experiment 1} & 0.003 $\pm$ 0.00 & 0.000 $\pm$ 0.00 & 0.008 $\pm$ 0.00 & 0.003 $\pm$ 0.00 \\
&\multicolumn{2}{l}{Human Experiment 2} & 0.037 $\pm$ 0.01   & 0.011 $\pm$ 0.01  & 0.058 $\pm$ 0.01 & 0.027 $\pm$ 0.01 \\
&\multicolumn{2}{l}{IBL-Equal} & 0.098 $\pm$ 0.03 & 0.016 $\pm$ 0.03 & 0.099 $\pm$ 0.02 & 0.024 $\pm$ 0.02\\
&\multicolumn{2}{l}{IBL-Exponential} & 0.095 $\pm$ 0.04 & 0.001 $\pm$ 0.04 & 0.096 $\pm$ 0.03 & 0.015 $\pm$ 0.03\\
&\multicolumn{2}{l}{IBL-TD} & 0.148 $\pm$ 0.01  & 0.002 $\pm$ 0.01 & 0.142 $\pm$ 0.02  & 0.011 $\pm$ 0.02\\ 
&\multicolumn{2}{l}{Q-Learning} & 0.161 $\pm$ 0.01  & 0.004 $\pm$ 0.01 & 0.150 $\pm$ 0.01  & 0.022 $\pm$ 0.01\\\bottomrule
\end{tabular}
}
\caption{Descriptive statistics (mean $\pm$ standard deviation) regarding \textbf{redundancy} and \textbf{immediate redundancy} during the first 10 episodes and last 10 playing episodes.}
\label{tab:redundancy_stat}
\end{table}

% \begin{figure}[!htbp]
% \centering
% \includegraphics[width=0.91\linewidth]{im_redundancy_stat.png}
% \caption{The average \textbf{immediate redundancy} rate in the first 10 and last 10 episodes.}
% \label{fig:immediate_redunancy_2}
% \end{figure}

% %---------------------------------
% %--------Linear Movement----------
% %---------------------------------
\subsection{Linear movement strategy}
\begin{figure}[!htbp]
\centering
\includegraphics[width=0.91\linewidth]{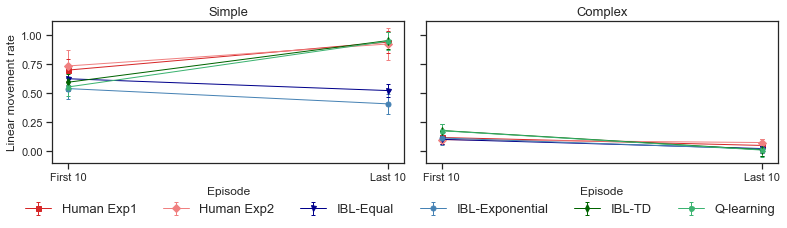}
\caption{The average \textbf{linear movement} rate in the first 10 and last 10 episodes.}
\label{fig:linear_movement_2}
\end{figure}

Table \ref{tab:anova_linear_movement} reports the decision complexity effects on executing linear movement strategy.
Table \ref{tab:tukey_linear_movement} compares the mean proportion of times following linear movement strategy among humans and the models.
Table~\ref{tab:linear_movement_stat} and Fig.~\ref{fig:linear_movement_2} show the average linear movement rate in the first and last 10 episodes.

\begin{table}[!htpb]
\resizebox{.8\textwidth}{!}{% 
\begin{tabular}{@{}llllrr@{}}
\toprule
 & Effect & DFn & DFd & F & p \\ \midrule
\multirow{3}{*}{Human Experiment 1} & complexity &  1 & 204 & 1139.70 & \textbf{0.000} \\
 & episode & 39 & 7956 & 4.46 & \textbf{0.000} \\
 & complexity:episode & 39 & 7956 & 16.80 & \textbf{0.000}  \\ \midrule
\multirow{3}{*}{Human Experiment 2} & complexity &  1 & 192 & 799.49 & \textbf{0.000}  \\
 & episode & 39 & 7488 & 6.70 & \textbf{0.000} \\
 & complexity:episode & 39 & 7488 & 10.50 & \textbf{0.000}  \\ \midrule
\multirow{3}{*}{IBL-Equal} & complexity & 1 & 124 & 93.70 & \textbf{0.000} \\
 & episode & 39 & 4836 & 4.36 & \textbf{0.000} \\
 & complexity:episode & 39 & 4836 & 0.96 & 0.549 \\ \midrule
\multirow{3}{*}{IBL-Exponential} & complexity & 1 & 124 & 52.78 & \textbf{0.000} \\
 & episode & 39 & 4836 & 7.24 & \textbf{0.000}  \\
 & complexity:episode & 39 & 4836 & 1.00 & 0.471  \\ \midrule
\multirow{3}{*}{IBL-TD} & complexity & 1 & 124 & 827.79 & \textbf{0.000} \\
 & episode & 39 & 4836 & 4.18 & \textbf{0.000} \\
 & complexity:episode & 39 & 4836 & 14.87 & \textbf{0.000}  \\ \midrule
\multirow{3}{*}{Q-Learning} & complexity & 1 & 124 & 889.60 & \textbf{0.000}  \\
 & episode & 39 & 4836 & 3.33 & \textbf{0.000}  \\
 & complexity:episode & 39 & 4836 & 16.54 & \textbf{0.000} \\ \bottomrule
\end{tabular}
}
\caption{ANOVA for the effect of decision complexity on the \textbf{linear movement strategy} of humans and the models.}
\label{tab:anova_linear_movement}
\end{table}

\begin{table}[!htpb]
\resizebox{.7\textwidth}{!}{% 
\begin{tabular}{@{}llllll@{}}
\toprule
\multicolumn{2}{l}{} & \multicolumn{2}{c}{Simple} & \multicolumn{2}{c}{Complex} \\ \cmidrule(l){3-6} 
Group 1 & Group 2 & \multicolumn{1}{c}{diff} & \multicolumn{1}{c}{p.adj} & \multicolumn{1}{c}{diff} & \multicolumn{1}{c}{p.adj} \\ \midrule
Human Experiment 2 & Human Experiment 1 & 0.001 & 1.000 & 0.012 & 0.289 \\
IBL-Equal & Human Experiment 1 & -0.292 & 0.000 & -0.027 & 0.000 \\
IBL-Exponential & Human Experiment 1 & -0.407 & 0.000 & -0.017 & 0.087 \\
IBL-TD & Human Experiment 1 & -0.102  & 0.000 & 0.028  & 0.000 \\
Q-learning & Human Experiment 1 & -0.121  & 0.000 & 0.028  & 0.000 \\
IBL-Equal & Human Experiment 2 & -0.293 & 0.000 & -0.040  & 0.000 \\
IBL-Exponential & Human Experiment 2 & -0.408 & 0.000 & -0.029 & 0.000   \\ 
IBL-TD & Human Experiment 2 & -0.103 & 0.000 & 0.016  & 0.169\\ 
Q-learning & Human Experiment 2 & -0.122  & 0.000  & 0.016  & 0.147\\ 
IBL-Exponential & IBL-Equal & -0.115  & 0.000  & 0.010 & 0.741 \\ 
IBL-TD & IBL-Equal & 0.190 & 0.000 & 0.055  & 0.000\\ 
Q-learning & IBL-Equal & 0.171 & 0.000  & 0.056 & 0.000 \\ 
IBL-TD & IBL-Exponential & 0.305 & 0.000 & 0.045  & 0.000\\ 
Q-learning & IBL-Exponential & 0.286  & 0.000 & 0.046 & 0.000 \\
Q-learning & IBL-TD & -0.019 & 0.550 & 0.000 & 1.000  \\ \bottomrule
\end{tabular}
}
\caption{ANOVA with post-hoc Tukey HSD for comparing the mean \textbf{linear movement} rate among humans and the models.}
\label{tab:tukey_linear_movement}
\end{table}

% %descriptive statistics
\begin{table}[!htpb]
\resizebox{.9\textwidth}{!}{% 
\begin{tabular}{@{}llllll@{}}
\toprule
\multicolumn{2}{l}{\multirow{2}{*}{Humans and Models}} & \multicolumn{2}{c}{Simple} & \multicolumn{2}{c}{Complex} \\ \cmidrule(l){3-6} 
\multicolumn{2}{l}{} & First 10 episodes & Last 10 episodes & First 10 episodes & Last 10 episodes \\ \midrule
\multicolumn{2}{l}{Human Experiment 1} & 0.698 $\pm$ 0.09 & 0.938 $\pm$ 0.09 & 0.119  $\pm$ 0.05  & 0.051 $\pm$ 0.05  \\
\multicolumn{2}{l}{Human Experiment 2} & 0.733 $\pm$ 0.14  & 0.922 $\pm$ 0.14  & 0.100 $\pm$ 0.03 & 0.076 $\pm$ 0.03 \\
\multicolumn{2}{l}{IBL-Equal} & 0.623 $\pm$ 0.05 & 0.522 $\pm$ 0.05 & 0.103 $\pm$ 0.04 & 0.024 $\pm$ 0.04 \\
\multicolumn{2}{l}{IBL-Exponential} & 0.539 $\pm$ 0.09 & 0.408 $\pm$ 0.09 & 0.114 $\pm$ 0.06 & 0.019 $\pm$ 0.06 \\
\multicolumn{2}{l}{IBL-TD} & 0.594 $\pm$ 0.07  & 0.952 $\pm$ 0.07 & 0.179 $\pm$ 0.05  &  0.015 $\pm$ 0.05\\ 
\multicolumn{2}{l}{Q-Learning} & 0.553 $\pm$ 0.08  & 0.947 $\pm$ 0.08 & 0.177 $\pm$ 0.06  &  0.011 $\pm$ 0.06 \\ \bottomrule
\end{tabular}
}
\caption{Descriptive statistics (mean $\pm$ standard deviation) regarding \textbf{linear movement strategy} during the first 10 and last 10 playing episodes.}
\label{tab:linear_movement_stat}
\end{table}

% %---------------------------------
% %-------------Coverage------------
% %---------------------------------
\subsection{Coverage}
\begin{figure}[!htbp]
\centering
\includegraphics[width=0.91\linewidth]{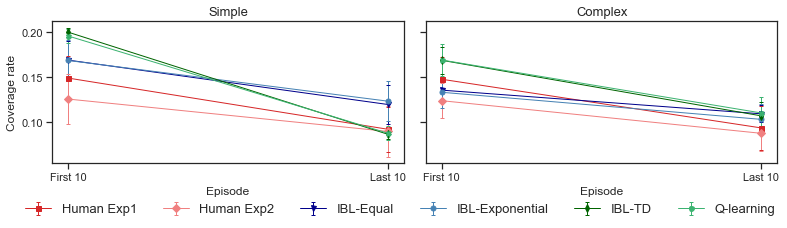}
\caption{The average percent of \textbf{coverage} in the first 10 and last 10 episodes.}
\label{fig:coverage_2}
\end{figure}

Table \ref{tab:anova_coverage} and \ref{tab:tukey_coverage} show the effects on the coverage of humans and the models.
Table~\ref{tab:coverage_stat} and Fig.~\ref{fig:coverage_2} display the mean percent of coverage in the first 10 and last 10 episodes.

\begin{table}[H]
\resizebox{.63\textwidth}{!}{% 
\begin{tabular}{@{}llllrr@{}}
\toprule
 & Effect & DFn & DFd & F & p \\ \midrule
\multirow{3}{*}{Human Experiment 1} & complexity & 1 & 204 & 0.02 & 0.885 \\
 & episode & 39 & 7956 & 29.68 & \textbf{0.000} \\
 & complexity:episode & 39 & 7956 & 0.68 & 0.938 \\ \midrule
\multirow{3}{*}{Human Experiment 2} & complexity & 1 & 192 & 0.33 & 0.568  \\
 & episode & 39.00 & 7488.00 & 23.47 & \textbf{0.000} \\
 & complexity:episode & 39.00 & 7488.00 & 1.53 & \textbf{0.019} \\ \midrule
\multirow{3}{*}{IBL-Equal} & complexity & 1 & 124 & 15.51 & \textbf{0.000} \\
 & episode & 39 & 4836 & 32.19 & \textbf{0.000} \\
 & complexity:episode & 39 & 4836 & 4.67 & \textbf{0.000} \\ \midrule
\multirow{3}{*}{IBL-Exponential} & complexity & 1 & 124 & 32.75 & \textbf{0.000} \\
 & episode & 39 & 4836 & 34.04 & \textbf{0.000} \\
 & complexity:episode & 39 & 4836 & 1.95 & \textbf{0.000} \\ \midrule
\multirow{3}{*}{IBL-TD} & complexity &  1 & 124 & 5.36 & \textbf{0.022}  \\
 & episode & 39 & 4836 & 64.53 & \textbf{0.000} \\
 & complexity:episode & 39 & 4836 & 7.81 & \textbf{0.000} \\ \midrule
\multirow{3}{*}{Q-Learning} & complexity & 1 & 124 & 5.85 & \textbf{0.017}  \\
 & episode & 39 & 4836 & 61.17 & \textbf{0.000} \\
 & complexity:episode &  39 & 4836 & 7.38 & \textbf{0.000}   \\ \bottomrule
\end{tabular}
}
\caption{ANOVA for the effect of decision complexity on the \textbf{coverage} of humans and the models.}
\label{tab:anova_coverage}
\end{table}

\begin{table}[!htbp]
\resizebox{.7\textwidth}{!}{% 
\begin{tabular}{@{}llllll@{}}
\toprule
\multicolumn{2}{l}{} & \multicolumn{2}{c}{Simple} & \multicolumn{2}{c}{Complex} \\ \cmidrule(l){3-6} 
Group 1 & Group 2 & \multicolumn{1}{c}{diff} & \multicolumn{1}{c}{p.adj} & \multicolumn{1}{c}{diff} & \multicolumn{1}{c}{p.adj} \\ \midrule
Human Experiment 2 & Human Experiment 1 &  -0.009  & 0.000  & -0.012 & 0.000  \\
IBL-Equal & Human Experiment 1 & 0.024 & 0.000  & 0.006 & 0.002 \\
IBL-Exponential & Human Experiment 1 & 0.027  & 0.000  & 0.000  & 1.000 \\
IBL-TD & Human Experiment 1 & 0.028  & 0.000  & 0.035  & 0.000 \\
Q-learning & Human Experiment 1 & 0.029 & 0.000 & 0.036  & 0.000 \\
IBL-Equal & Human Experiment 2  & 0.032  & 0.000  & 0.018  & 0.000 \\
IBL-Exponential & Human Experiment 2 & 0.036  & 0.000  & 0.012  & 0.000   \\ 
IBL-TD & Human Experiment 2 & 0.037  & 0.000 & 0.047  & 0.000\\ 
Q-learning & Human Experiment 2  & 0.037  & 0.000  & 0.048  & 0.000  \\ 
IBL-Exponential & IBL-Equal & 0.003  & 0.266  & -0.006  & 0.015 \\ 
IBL-TD & IBL-Equal & 0.004  & 0.075 & 0.029  & 0.000 \\ 
Q-learning & IBL-Equal & 0.005 & 0.014   & 0.030  & 0.000 \\ 
IBL-TD & IBL-Exponential & 0.001  & 0.993  & 0.035  & 0.000\\ 
Q-learning & IBL-Exponential & 0.002 & 0.873  & 0.036  & 0.000  \\
Q-learning & IBL-TD & 0.001 & 0.993  & 0.001  & 0.993  \\ \bottomrule
\end{tabular}
}
\caption{ANOVA with post-hoc Tukey HSD for comparing the mean \textbf{coverage} rate among humans and the models.}
\label{tab:tukey_coverage}
\end{table}

% %descriptive statistics
\begin{table}[!htbp]
\resizebox{.9\textwidth}{!}{% 
\begin{tabular}{@{}llllll@{}}
\toprule
\multicolumn{2}{l}{\multirow{2}{*}{Humans and Models}} & \multicolumn{2}{c}{Simple} & \multicolumn{2}{c}{Complex} \\ \cmidrule(l){3-6} 
\multicolumn{2}{l}{} & First 10 episodes & Last 10 episodes & First 10 episodes & Last 10 episodes \\ \midrule
\multicolumn{2}{l}{Human Experiment 1} & 0.149 $\pm$ 0.02 & 0.092 $\pm$ 0.02 & 0.148  $\pm$ 0.02  & 0.094 $\pm$ 0.02  \\
\multicolumn{2}{l}{Human Experiment 2} & 0.126 $\pm$ 0.03  & 0.089 $\pm$ 0.03  & 0.124 $\pm$ 0.02 & 0.088 $\pm$ 0.02 \\
\multicolumn{2}{l}{IBL-Equal} & 0.169 $\pm$ 0.02 & 0.119 $\pm$ 0.02 & 0.136 $\pm$ 0.01 & 0.109 $\pm$ 0.01 \\
\multicolumn{2}{l}{IBL-Exponential} & 0.168 $\pm$ 0.02 & 0.123 $\pm$ 0.02 & 0.133 $\pm$ 0.02 & 0.103 $\pm$ 0.02 \\
\multicolumn{2}{l}{IBL-TD} & 0.199 $\pm$ 0.00  & 0.087 $\pm$ 0.00 & 0.169 $\pm$ 0.01  &  0.107 $\pm$ 0.01\\
\multicolumn{2}{l}{Q-Learning} & 0.196 $\pm$ 0.01  & 0.088 $\pm$ 0.01 & 0.169 $\pm$ 0.02  &  0.110 $\pm$ 0.02\\ \bottomrule
\end{tabular}
}
\caption{Descriptive statistics (mean $\pm$ standard deviation) regarding \textbf{coverage} during the first 10 and last 10 playing episodes.}
\label{tab:coverage_stat}
\end{table}

% %---------------------------------
% %------Closest distractor---------
% %---------------------------------
\subsection{Closest distractor}
\begin{figure}[H]
\centering
\includegraphics[width=0.91\linewidth]{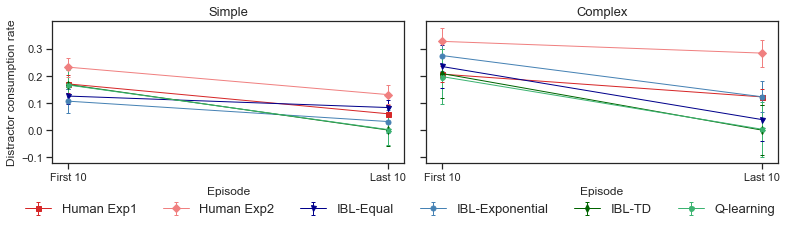}
\caption{The \textbf{closest distractor} consumption rate in the first 10 and last 10 episodes.}
\label{fig:distractor_2}
\end{figure}

Table \ref{tab:anova_distractor} and \ref{tab:tukey_distractor} show the effects on the rate of getting the closest distractor.
Table~\ref{tab:closest_distractor_stat} and Fig.~\ref{fig:distractor_2} report the mean distractor consumption in the first and last 10 episodes.

\begin{table}[!htbp]
\resizebox{.6\textwidth}{!}{% 
\begin{tabular}{@{}llllll@{}}
\toprule
 & Effect & DFn & DFd & F & p \\ \midrule
\multirow{3}{*}{Human Experiment 1} & complexity &  1 & 204 & 3.76 & 0.054\\
 & episode &  39 & 7956 & 5.63 & \textbf{0.000} \\
 & complexity:episode &  39 & 7956 & 0.80 & 0.807 \\ \midrule
\multirow{3}{*}{Human Experiment 2} & complexity &  1 & 192 & 6.44 & \textbf{0.012}   \\
 & episode & 39 & 7488 & 3.09 & \textbf{0.000} \\
 & complexity:episode &  39 & 7488 & 1.16 & 0.225 \\ \midrule
\multirow{3}{*}{IBL-Equal} & complexity & 1 & 124 & 0.24 & 0.625 \\
 & episode & 39 & 4836 & 5.97 & \textbf{0.000} \\
 & complexity:episode & 39 & 4836 & 3.51 & \textbf{0.000} \\ \midrule
\multirow{3}{*}{IBL-Exponential} & complexity & 1 & 124 & 9.15 & \textbf{0.003} \\
 & episode & 39 & 4836 & 6.82 & \textbf{0.000} \\
 & complexity:episode & 39 & 4836 & 1.18 & 0.211 \\ \midrule
\multirow{3}{*}{IBL-TD} & complexity & 1 & 124 & 0.02 & 0.890 \\
 & episode &  39 & 4836 & 16.59 & \textbf{0.000} \\
 & complexity:episode & 39 & 4836 & 3.02 & \textbf{0.000} \\ \midrule
\multirow{3}{*}{Q-Learning} & complexity & 1 & 124 & 0.15 & 0.699 \\
 & episode & 39 & 4836 & 17.88 & \textbf{0.000} \\
 & complexity:episode & 39 & 4836 & 1.64 & \textbf{0.007} \\ \bottomrule
\end{tabular}
}
\caption{ANOVA for the effect of decision complexity on the \textbf{closest distractor consumption} of humans and the models.}
\label{tab:anova_distractor}
\end{table}

\begin{table}[!htbp]
\resizebox{.7\textwidth}{!}{% 
\begin{tabular}{@{}llllll@{}}
\toprule
\multicolumn{2}{l}{} & \multicolumn{2}{c}{Simple} & \multicolumn{2}{c}{Complex} \\ \cmidrule(l){3-6} 
Group 1 & Group 2 & \multicolumn{1}{c}{diff} & \multicolumn{1}{c}{p.adj} & \multicolumn{1}{c}{diff} & \multicolumn{1}{c}{p.adj} \\ \midrule
Human Experiment 2 & Human Experiment 1 & 0.071 & 0.000 & 0.127  & 0.000 \\
IBL-Equal & Human Experiment 1 & -0.007 & 0.947  & -0.052  & 0.000 \\
IBL-Exponential & Human Experiment 1 & -0.044  & 0.000  & 0.015 & 0.583 \\
IBL-TD & Human Experiment 1 & -0.047  & 0.000 & -0.109  & 0.000  \\
Q-learning & Human Experiment 1 & -0.048  & 0.000  & -0.109  & 0.000  \\
IBL-Equal & Human Experiment 2 & -0.078  & 0.000   & -0.179  & 0.000 \\
IBL-Exponential & Human Experiment 2 & -0.115 & 0.000 & -0.112  & 0.000   \\ 
IBL-TD & Human Experiment 2 & -0.117  & 0.000  & -0.236  & 0.000\\ 
Q-learning & Human Experiment 2 & -0.119  & 0.000  & -0.236  & 0.000\\ 
IBL-Exponential & IBL-Equal & -0.038  & 0.000   & 0.067  & 0.000 \\ 
IBL-TD & IBL-Equal & -0.040  & 0.000  & -0.057  & 0.000 \\ 
Q-learning & IBL-Equal & -0.042  & 0.000  & -0.057 & 0.000 \\ 
IBL-TD & IBL-Exponential & -0.002  & 1.000 & -0.124  & 0.000 \\ 
Q-learning & IBL-Exponential  & -0.004  & 0.995 & -0.124  & 0.000  \\
Q-learning & IBL-TD & -0.002 & 1.000  & 0.000  & 1.000  \\ \bottomrule
\end{tabular}
}
\caption{ANOVA with post-hoc Tukey HSD for comparing the mean \textbf{closest distractor consumption} rate among humans and the models.}
\label{tab:tukey_distractor}
\end{table}

\begin{table}[!htbp]
\resizebox{.9\textwidth}{!}{% 
\begin{tabular}{@{}llllll@{}}
\toprule
\multicolumn{2}{l}{\multirow{2}{*}{Humans and Models}} & \multicolumn{2}{c}{Simple} & \multicolumn{2}{c}{Complex} \\ \cmidrule(l){3-6} 
\multicolumn{2}{l}{} & First 10 episodes & Last 10 episodes & First 10 episodes & Last 10 episodes\\ \midrule
\multicolumn{2}{l}{Human Experiment 1} & 0.169 $\pm$ 0.03 & 0.059 $\pm$ 0.03 & 0.206 $\pm$ 0.03 & 0.122 $\pm$ 0.03 \\
\multicolumn{2}{l}{Human Experiment 2} & 0.231 $\pm$ 0.03   & 0.130 $\pm$ 0.03  & 0.326 $\pm$ 0.05 & 0.283 $\pm$ 0.05 \\
\multicolumn{2}{l}{IBL-Equal} & 0.125 $\pm$ 0.03 & 0.083 $\pm$ 0.03 & 0.234 $\pm$ 0.08 & 0.039 $\pm$ 0.08\\
\multicolumn{2}{l}{IBL-Exponential} & 0.106 $\pm$ 0.04 & 0.031 $\pm$ 0.04 & 0.274 $\pm$ 0.06 & 0.123 $\pm$ 0.06\\
\multicolumn{2}{l}{IBL-TD} & 0.167 $\pm$ 0.06  & 0.000 $\pm$ 0.06 & 0.208 $\pm$ 0.09  & 0.000 $\pm$ 0.09\\ 
\multicolumn{2}{l}{Q-Learning} & 0.166 $\pm$ 0.06  & 0.001 $\pm$ 0.06 & 0.197 $\pm$ 0.10  & 0.003 $\pm$ 0.10\\\bottomrule
\end{tabular}
}
\caption{Descriptive statistics (mean $\pm$ standard deviation) regarding \textbf{closest distractor consumption} during the first 10 and last 10 playing episodes.}
\label{tab:closest_distractor_stat}
\end{table}

% \end{appendices}

\end{document}